\documentclass{article}

\usepackage{PRIMEarxiv}

\usepackage{geometry}
\usepackage{longtable}
\usepackage{threeparttable}
\usepackage{subcaption}
\usepackage{amsmath}
\usepackage{float}
\usepackage{comment}
\usepackage[utf8]{inputenc} % allow utf-8 input
\usepackage[T1]{fontenc}    % use 8-bit T1 fonts
\usepackage{hyperref}       % hyperlinks
\usepackage{url}            % simple URL typesetting
\usepackage{booktabs}       % professional-quality tables
\usepackage{amsfonts}       % blackboard math symbols
\usepackage{nicefrac}       % compact symbols for 1/2, etc.
\usepackage{microtype}      % microtypography
\usepackage{lipsum}
\usepackage{fancyhdr}       % header
\usepackage{graphicx}       % graphics
\usepackage{tabularx}
\usepackage{authblk}
\usepackage{pdflscape}
\usepackage{multirow} 
\usepackage{colortbl}
\usepackage{xcolor}
\usepackage{array}
\usepackage{orcidlink}
\usepackage{silence}
\graphicspath{{graphs/}}

\title{\textbf{Reliability, Resilience and Human Factors Engineering for Trustworthy AI Systems}}

\author{
\begin{center}
    \begin{minipage}{0.3\textwidth}
        \centering
        \textbf{Saurabh Mishra} \\
        \vspace{-1em}
        \textit{Taiyō.AI, The Brookings Institution, OECD.AI}
    \end{minipage}
    \hfill
    \begin{minipage}{0.3\textwidth}
        \centering
        \textbf{Anand Rao} \\
        \vspace{-1em}
        \textit{Carnegie Mellon University}
    \end{minipage}
    \hfill
    \begin{minipage}{0.3\textwidth}
        \centering
        \textbf{Ramayya Krishnan} \\
        \vspace{-1em}
        \textit{Carnegie Mellon University}
    \end{minipage}
    \vspace{2em} % Adds vertical space between rows
    \\
    \begin{minipage}{0.3\textwidth}
        \centering
        \textbf{Bilal Ayyub} \\
        \vspace{0.1em}
        \textit{University of Maryland, \newline College Park}
        \
    \end{minipage}
    \hfill
    \begin{minipage}{0.3\textwidth}
        \centering
        \textbf{Amin Aria} \\
        \vspace{0.1em}
        \textit{ARETUM Labs}
    \end{minipage}
    \hfill
    \begin{minipage}{0.3\textwidth}
        \centering
        \textbf{Enrico Zio} \\
        \vspace{0.1em}
        \textit{Mines Paris PSL University \newline Energy Department, Politecnico di Milano}
    \end{minipage}
\end{center}
}

\begin{document}
\raggedbottom 
\maketitle

\begin{abstract}
As AI systems become integral to critical operations across industries and services, ensuring their reliability and safety is essential. We offer a framework that integrates established reliability and resilience engineering principles into AI systems. By applying traditional metrics such as failure rate and Mean Time Between Failures (MTBF) along with resilience engineering and human reliability analysis, we propose an integrate framework to manage AI system performance, and prevent or efficiently recover from failures. Our work adapts classical engineering methods to AI systems and outlines a research agenda for future technical studies. We apply our framework to a real-world AI system, using system status data from platforms such as openAI, to demonstrate its practical applicability. This framework aligns with emerging global standards and regulatory frameworks, providing a methodology to enhance the trustworthiness of AI systems. Our aim is to guide policy, regulation, and the development of reliable, safe, and adaptable AI technologies capable of consistent performance in real-world environments.
\end{abstract}

% keywords can be removed

\section{Introduction}

Ensuring the safety of AI systems has become a central concern as these systems are increasingly integrated into our daily lives. At the heart of this discourse is reliability engineering, a field that has guided the reliability of complex engineering systems for decades. Reliability engineering involves applying scientific know-how to ensure that systems perform their intended functions without failure over a specified period. In this paper, following system reliability conventions, failure refers to any deviation from expected performance. We propose a framework that integrates Reliability Engineering, Resilience Engineering, and Human Factors Engineering, underpinned by Prognostics and Health Management (PHM), to enhance the trustworthiness of AI systems. In our framework, reliability engineering addresses pre-deployment considerations to prevent failures once the system is operational while resilience engineering focuses on post-deployment to ensure the system's ability to recover from failures. Human factors are also integral throughout both phases. Central to our approach is the Human-Centric AI Reliability Model (HC-AIRM), which emphasizes the role of human factors in both reliability and resilience engineering, ensuring that human interactions are considered throughout the AI system lifecycle (see Section \ref{sec:HCAIRM}).

An AI system is composed of multiple subsystems, including Data, Model, Computing Infrastructure, Code and Software, and Human interaction. Each subsystem contains components that contribute to the overall functionality of the AI system. Understanding the interplay between these subsystems and their components is essential for PHM of AI lifecycle and ensuring system reliability and resilience. In the meantime, it should be recognized that technical reliability and human reliability are tightly intertwined and cannot be considered separately.

A recent real-world example illustrates the critical need for integrating human factors into technical reliability. In January 2024, news outlets reported on the potential political influence of public figures like Taylor Swift, highlighting her sway over young social media followers \cite{NYTimes2024} through rapid dissemination of deepfake images across social networks like X (formerly known as Twitter) \cite{FT2024}. These deepfakes raised immediate concerns about the misuse and weaponization of AI technologies, prompting swift actions such as restricting specific search queries, as part of post-deployment resilience, to mitigate the harm.

AI systems are diverse, with diverse components/modules each associated with specific failure modes that are domain- and context- specific. These can be broadly categorized into intentional (adversarial) and unintentional (design flaws, human errors) failures \cite{kumar2024failure}. Intentional failures occur when external actors exploit vulnerabilities through tactics like model poisoning or data manipulation. For instance, in  autonomous vehicles adversarial attackers can alter road signs or display adversarial patches on moving vehicles to deceive object detection models, leading to incorrect decision-making processes that exploit model's perception capabilities and data vulnerabilities \cite{houfaultanalysis, chahe2023dynamic}. Unintentional failures generally stem from design flaws, incomplete testing, or distributional shifts during the AI lifecycle. In the field of radiology, as an example, AI systems interpreting medical images can produce false diagnoses or miss critical conditions due to inadequate training data, biases in the dataset, or limitations in the model architecture \cite{yu2023evaluating}. Such errors not only affect the reliability of the AI system but also have the potential to negatively influence human decision-making.  Users often expect lower error rates for AI systems compared to human counterparts, underscoring the demand for higher reliability in AI applications \cite{lenskjold2023should}. As such, AI systems must be designed with a deep understanding of both technical limitations and human expectations.

Incidents and examples explain above underscores how quickly AI can be exploited in unforeseen ways or cause catastrophic failures, highlighting the importance of considering misuse and malfunction at the design stage, which can be addressed using the Human-centric PHM approach proposed in this paper. By integrating human factors into the design process, we can implement remedial solutions or fail-safe mechanisms from the outset to prevent such failures. For instance, incorporating watermarking techniques into generative AI models can help identify synthetic content and deter misuse, exemplifying how pre-deployment reliability measures can enhance post-deployment resilience \cite{Dathathri2024}.

Analogous to how IKEA designs its furniture components to prevent incorrect assembly—where parts are shaped so that they cannot be put together improperly—we advocate for and propose a human-centric PHM framework that uses reliability engineering best practices to design AI systems with built-in safeguards. By anticipating worst-case scenarios and ensuring that both components and higher-level systems can handle them acceptably, we move towards trustworthy AI systems capable of consistent performance in real-world environments.

Introducing metrics from reliability engineering, such as failure rate and Mean Time Between Failures (MTBF) offers a structured approach to measuring and managing AI system performance over time. By understanding these failures within a reliability engineering framework and incorporating cost-benefit analysis, we can propose targeted interventions—such as improved algorithms or better system controls—that balance the costs and benefits of ensuring AI system safety \cite{groth2012,jan2022}. Defining what constitutes an AI failure across sectors or tasks in finance, healthcare, or critical infrastructure is domain-specific and is key to developing a comprehensive approach to AI reliability and resilience. Moreover, the societal impact—whether measured in economic terms, number of people affected, or critical infrastructure compromised—varies widely, reinforcing the importance of applying these metrics across various domains.

\begin{figure}[htbp]
    \centering
    \begin{subfigure}[b]{0.5\textwidth}
        \centering
        \includegraphics[width=\textwidth]{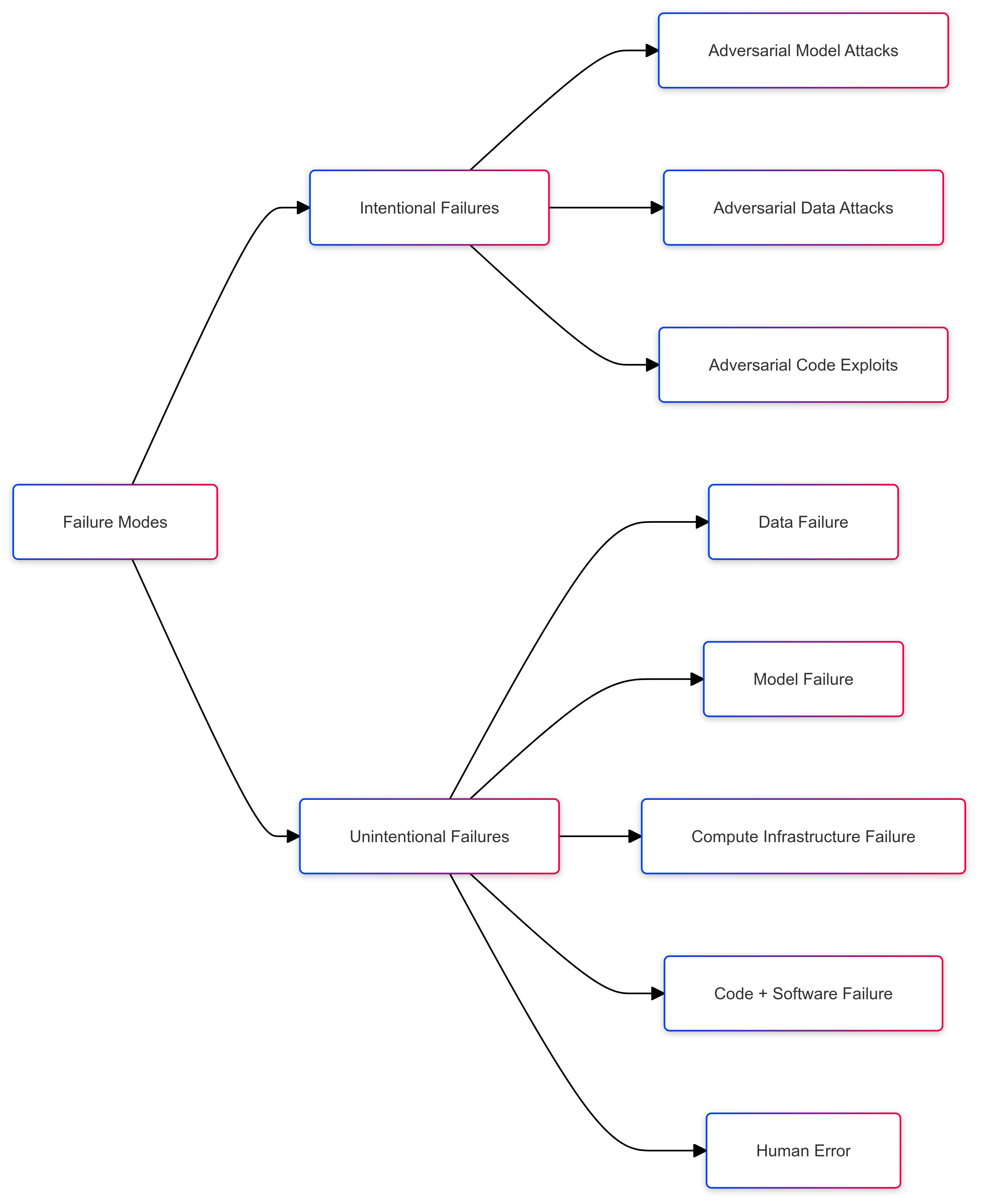}
        \caption{High-Level Overview of AI Failure Modes and Types}
        \label{fig:fig1a}
    \end{subfigure}
    \vspace{1em} % Add space between the subfigures
    \begin{subfigure}[b]{0.8\textwidth}
        \centering
        \includegraphics[width=0.9\textwidth]{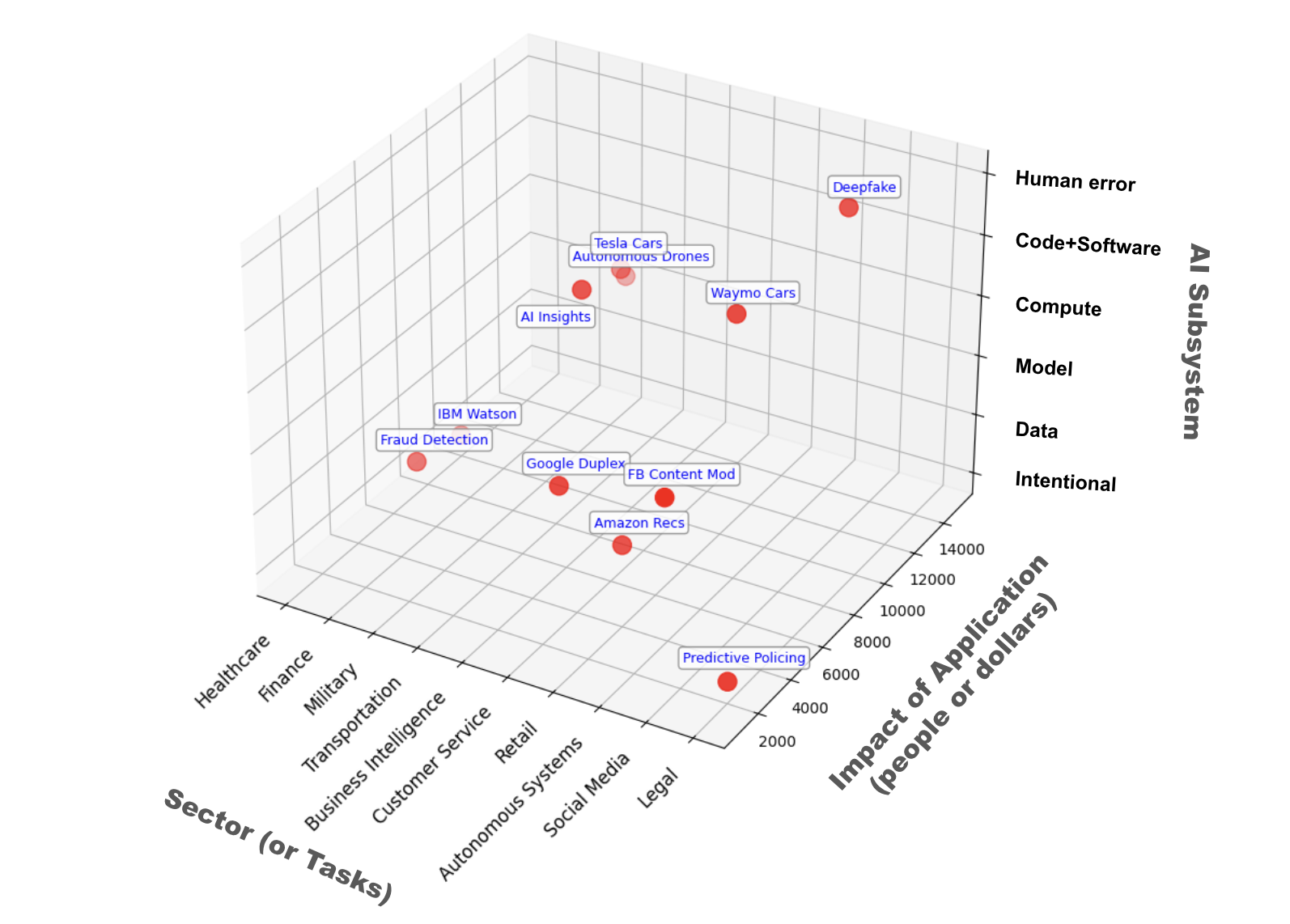}
        \caption{Hypothetical 3D Matrix of Sectors (Tasks), Impacts, and Failure Modes}
        \label{fig:fig1b}
    \end{subfigure}
    \caption{Illustrating AI Failure Modes: High-Level Taxonomy and Relating Sector-Impact-Failure Modes}
    \label{fig:main_figure}
\end{figure}

Figure~\ref{fig:main_figure} illustrates a high-level taxonomy of AI failure modes and relates these modes to various sectors, impacts, and tasks, emphasizing the dynamic interlinkage between application-specific sectors and the critical need for resilient AI systems across industries. By visualizing these connections, we can better understand the trade-offs and areas requiring focused interventions to ensure reliable performance. 

While reliability is the cornerstone of system performance, trustworthiness, which is critical is AI systems as it influences user confidence and the societal acceptance of AI technologies, encompasses a broader spectrum, including aspects like transparency, explainability, fairness and security. Trustworthiness . Unlike traditional systems, where trust is often built through consistent performance, AI systems must also demonstrate their ability to adapt, learn, and improve while maintaining reliability \cite{DIAZRODRIGUEZ2023101896, EU2019, HAL2024}. 

Considering the dynamic and complex environment and architecture of AI System and the need for trustworthiness, reliability of such systems should not only refer to the consistent performance over time but also account for the system's ability to handle new data, adapt to changing environments, and maintain performance under varying conditions. In the context of reliability, AI systems are considered \textbf{``better than new''} repairable systems --- systems that have varied version updates with the objective that new releases outperform prior versions. AI systems are dynamic, meaning they can  ``age'' differently compared to traditional systems. Traditional systems may degrade slowly, but AI models can experience rapid ``aging'' through model drift, where shifts in data distributions lead to decreased performance \cite{barassi2020}. For example, a sentiment analysis model trained on a particular social media platform may misclassify sentiments if slang or language use evolves over time. Addressing this requires continuous adaptation and retraining \cite{almog2024AI, collins2023uncertainty}. AI reliability must consider these factors, making the concept more complex and intertwined with continuous learning and adaptation \cite{wu2024continuallearninglargelanguage, donakanti2024reimaginingselfadaptationagelarge}. Additionally, understanding the reliability of AI systems is crucial for calculating their Return on Investment (ROI), given the financial implications of both Type 1 and Type 2 errors in varied business contexts \cite{solvingAI2023}. 

To address the concerned discussed above, the proposed PHM framework of this paper considers system architecture and interplays through systematic review of the AI system using human-centric reliability and resilience engineering best practices. Additionally, it account for trustworthiness through insightful and holistic integration of traditional reliability and resilience engineering methodologies with AI systems. Doing so, the presented PHM framework tries to redefine reliability for AI systems and measure trustworthiness through ensuring both consistent reliability and fostering user confidence through transparent, ethical, and resilient AI practices. Thus,  AI Reliability in the proposed framework includes guidelines for building models that can detect and adapt to data drifts and trends, reducing the likelihood of errors and providing the ability to calculate the ROI given the financial implications in varied business contexts \cite{solvingAI2023}. 

Type I (false positive) and Type II (false negative) errors occur when the system either incorrectly identifies or misses an event, such as a medical diagnosis or a loan approval \cite{almog2024AI}. Both errors can stem from machine predictions or human interactions, especially when AI systems are used as recommendation engines \cite{collins2023uncertainty}. A finance example would be approval of an unqualified applicant (false positive) or rejection of a qualified one (false negative). While preventing these errors requires adaptive reliability engineering, managing them requires resilience—ensuring systems can recover from disruptions—and human factors engineering, which focuses on how humans interact with the system, including designing interfaces that help users understand and correct AI outputs \cite{barassi2020, almog2024AI}. The proposed PHM framework for AI systems integrate these layers by considering not only the model's performance but also the human elements of uncertainty, decision-making, and error handling. This holistic approach ensures robust performance, particularly in critical sectors like healthcare, finance, and autonomous systems, where reliability and resilience are paramount for safety and trust. 

In sum, this study presents a conceptual framework for improving safety of AI systems through the integration of human-centric reliability and resilience engineering principles, underpinned by PHM. Central to our methodology are concepts like the ``bathtub curve'', failure rate, MTBF, probabilistic risk, and resilient failure recovery, which we leverage to proactively manage AI system failures. Following a system level review, we focus on failure modes and PHM at the subsystem and component levels, emphasizing the importance of addressing failures in different subsystems of an AI system upon the release of new versions. By applying PHM methodologies at these granular levels, we aim to enhance both pre-deployment reliability and post-deployment resilience within the AI system lifecycle. 

Our analysis using the proposed framework also acknowledges the importance and urgency of emerging frameworks such as the
\href{https://www.nist.gov/itl/ai-risk-management-framework}{NIST AI Risk Management Framework}, \href{https://www.pdpc.gov.sg/help-and-resources/2020/01/model-ai-governance-framework}{Singapore’s Model Governance Framework} \& \href{https://fpf.org/blog/ai-verify-singapores-ai-governance-testing-initiative-explained/}{AI Verify}, the \href{https://eur-lex.europa.eu/legal-content/EN/TXT/?uri=celex\%3A52021PC0206}{EU AI Act}, \href{https://www.whitehouse.gov/ostp/ai-bill-of-rights/}{Blueprint for an AI Bill of Rights}, \href{https://www.un.org/en/ai-advisory-body}{United Nations AI Advisory Body}, \href{https://crfm.stanford.edu/fmti/}{Foundational Model Transparency Index}, \href{https://oecd.ai/en/catalogue/metrics}{the OECD's catalogue of tools and metrics}, all underlining the relevance and urgency of such approaches. Economic metrics, such as costs, revenues, and error rates, are pivotal in AI system reliability engineering, influencing financial outcomes and guiding policy and insurance frameworks \cite{jan2022}.

Incorporating PHM into AI systems provides a critical safety layer by enabling proactive maintenance and anomaly detection through predictive analytics. This continuous monitoring approach allows for early identification of issues, preventing minor faults from escalating into catastrophic failures \cite{heng2009rotating, si2011remaining, ramasso2014performance}. By applying PHM methodologies—originally developed for machinery monitoring—to AI, we can ensure ongoing assessments of model accuracy, data integrity, and infrastructure stability, akin to scheduling predictive maintenance in engineering. This integration is especially vital given the EU AI Act's emphasis on transparency and regulatory compliance, further underscoring the need for robust monitoring frameworks \cite{li2023chatgptlikelargescalefoundationmodels, compl-ai}. The EU AI Act, as detailed in the COMPL-AI Framework \cite{compl-ai} (\href{https://compl-ai.org/evaluations/gpt-4-1106-preview}{compl-ai.org}), highlights the necessity for actionable technical interpretations and benchmarking to align AI development with regulatory standards. Furthermore, resources like the AI Incident Database \cite{feffer2023aiid} and the AI Risk Repository \cite{slattery2023airisk} emphasize the importance of understanding AI harms and managing systemic risks through comprehensive evaluation frameworks.

The rest of the paper is organized as follows. Section~\ref{framework} introduces our systematic framework for PHM of AI systems, providing an overview of the AI lifecycle and how it integrates PHM concepts and principles for reliability and resilience, with specific attention to AI-specific failure modes in Subsection~\ref{failuremodes}. Section~\ref{reliability} delves into the \textit{AI System Reliability: Key Metrics and Engineering Approaches}, establishing the foundational metrics essential for assessing reliability in AI systems. This is complemented by Section~\ref{resilience}, which explores resilience engineering for AI Systems, focusing on mechanisms to enhance system robustness and adaptability under diverse operational conditions. Section~\ref{humanfactors} integrates human factors into AI reliability and resilience, emphasizes the importance of human-factors analysis in AI systems and  introduces the \textit{Human-Centric AI Reliability Model} to underscore the role of human interaction and oversight in maintaining AI system resilience. A practical application of the proposed framework is presented in Section~\ref{casestudy}, where a case study demonstrates real-world application. Section~\ref{conclusion} concludes with a summary of findings and future research directions. The \textit{Technical Appendix} provides additional depth on essential reliability engineering methods relevant to AI.

\section{Systematic Framework for PHM of AI Systems}
\label{framework}

The technical performance of AI systems is both diverse and evolving, encompassing various performance metrics across different tasks such as object detection, language modeling, and image recognition. Notable benchmarks, such as the aggregations conducted by Stanford AI Index's technical performance section \cite{maslej2024artificial}, and platforms like \href{https://paperswithcode.com/sota}{State-of-the-Art Papers with Code}, which maintains a free and open resource of Machine Learning papers, accompanying code, datasets, evaluation methods, and tables, provide systematic evaluations across models,capturing the nuanced performance criteria for different AI applications.

Reliability metrics in machine learning (ML) and deep learning (DL) involve concepts related to traditional performance measures such as precision, recall, and F-scores, often mapped to statistical Type I and Type II errors. These metrics play a critical role in improving MTBF, crucial indicators for robust model deployment. Prior to production deployment, models undergo rigorous reliability testing, a practice aligned with software testing but specifically adapted for AI model behavior under diverse conditions. This establishes an upfront focus on model reliability during testing and development stages, helping distinguish AI reliability from traditional software reliability frameworks.

To address the full spectrum of trustworthiness requirements, the \textit{OECD.AI Catalogue of Tools \& Metrics for Trustworthy AI} (Table \ref{oecd}) outlines key principles such as transparency, explainability, and robustness, mapping specific metrics to various AI actors, including data scientists, developers, and system integrators. These principles align well with our systematic framework, as they guide AI reliability and resilience by establishing performance standards that consider both technical and human-centric factors.

AI reliability is closely related to software reliability but introduces unique complexities due to the dynamic operating environment in which AI models function. Traditional software reliability often relies on Software Reliability Growth Models (SRGM) based on Non-Homogeneous Poisson Processes (NHPP), such as the S-shaped Gompertz model or the concave Weibull model, as discussed in classic studies \cite{wood, rotella}. Unlike traditional software, where the environment is relatively stable, AI model performance significantly depends on environmental factors, resulting in inherent, sometimes irreducible errors. For example, dual poisson processes has been used to model AI reliability, incorporating both fixed and decreasing failure rates, a framework extendable to more complex AI reliability challenges \cite{bastani}.

Recent developments in software reliability, such as PHM and NHPP-based models \cite{2017_Song}, have begun to adapt to AI reliability. For instance, studies (\cite{merkel2018softwarereliabilitygrowthmodels, hong2021reliabilityanalysisartificialintelligence}) applied SRGMs to analyze AI-related datasets, such as California AV disengagement data, while \cite{nafreen} developed models with bathtub-shaped fault intensity functions to balance model complexity and predictive accuracy in AI reliability analysis. These advancements in reliability modeling for AI extend traditional software reliability frameworks, accommodating AI's sensitivity to changing environments and non-removable intrinsic errors.

Reliability and resilience assessments of AI systems are traditionally conducted in corporate environments by product managers and internal governance teams. However, this paper seeks to codify these reliability concepts systematically for PHM of AI systems, establishing a framework akin to that of software reliability engineering. By consolidating these practices, this work aims to provide a structured approach to AI reliability and resilience engineering, addressing traditional software reliability assumptions and extending them to encompass the unique challenges posed by AI systems.

\begin{table}[htbp]
\centering
\caption{Metrics for Trustworthy AI Mapped by OECD AI Principle and Target User\\ \textit{Source: Adapted from OECD.AI Catalogue of Tools \& Metrics for Trustworthy AI}}
\begin{tabular}{|l|l|p{8cm}|}
\hline
\textbf{OECD AI Principle}            & \textbf{Target User}       & \textbf{Metrics}                                             \\ \hline
Transparency and Explainability       & Data Scientist             & SHAP, LIME, SAFE, LFISS                                       \\ \hline
Transparency and Explainability       & Developer                  & SHAP, Local Feature Importance Spread                         \\ \hline
Transparency and Explainability       & System Integrator          & LFISS, SAFE, CLIPSBERTScore                                   \\ \hline
Safety                                & Data Scientist             & False Acceptance Rate, False Rejection Rate                   \\ \hline
Safety                                & Developer                  & Structural Similarity Index, FAR                              \\ \hline
Safety                                & System Integrator          & Multi-Object Tracking Accuracy, FAR                           \\ \hline
Robustness and Digital Security       & Data Scientist             & Stability, Anonymity Set Size                                 \\ \hline
Robustness and Digital Security       & Developer                  & Tree Edit Distance, Variable Importance                       \\ \hline
Robustness and Digital Security       & System Integrator          & Structural Similarity Index, OOD Generalization               \\ \hline
Privacy and Data Governance           & Data Scientist             & Anonymity Set Size, Conditional Entropy                       \\ \hline
Privacy and Data Governance           & Developer                  & Amount of Leaked Information, Anonymity Set Size              \\ \hline
Privacy and Data Governance           & System Integrator          & Kendall Rank, Anonymity Set Size                              \\ \hline
Performance                           & Data Scientist             & Precision, Recall, F-score                                    \\ \hline
Performance                           & Developer                  & BLEU, ROUGE, MSE                                             \\ \hline
Performance                           & System Integrator          & Perplexity, MAE, Matthews Correlation                         \\ \hline
Fairness                              & Data Scientist             & Statistical Parity, Equal Opportunity                         \\ \hline
Fairness                              & Developer                  & Equality of Opportunity, Data Shapley                         \\ \hline
Fairness                              & System Integrator          & Rank-Aware Divergence, SPD                                    \\ \hline
Accountability                              &           &   Technical Metrics unidentified                                  \\ \hline
Human Well-Being                              &           &    Technical Metrics unidentified                                   \\ \hline
\end{tabular}
\label{oecd}
\end{table}

\subsection{Reliability-Resilience Framework for AI System Lifecycle} 
\label{framework2}

AI systems are complex systems that one can break down into distinct \textbf{`subsystems'} composed of \textbf{`components'} (or modules), each having a specific role in contributing to the system’s overall functionality. These subsystems include \textbf{data}, \textbf{models}, \textbf{compute infrastructure (cloud)}, \textbf{code+software} and \textbf{human}. Human errors can occur at multiple stages — during development (developer), deployment (deployer) or in production (end-user or DevOps engineer).

\begin{figure}[htbp]
    \centering
    \includegraphics[width=1\textwidth]{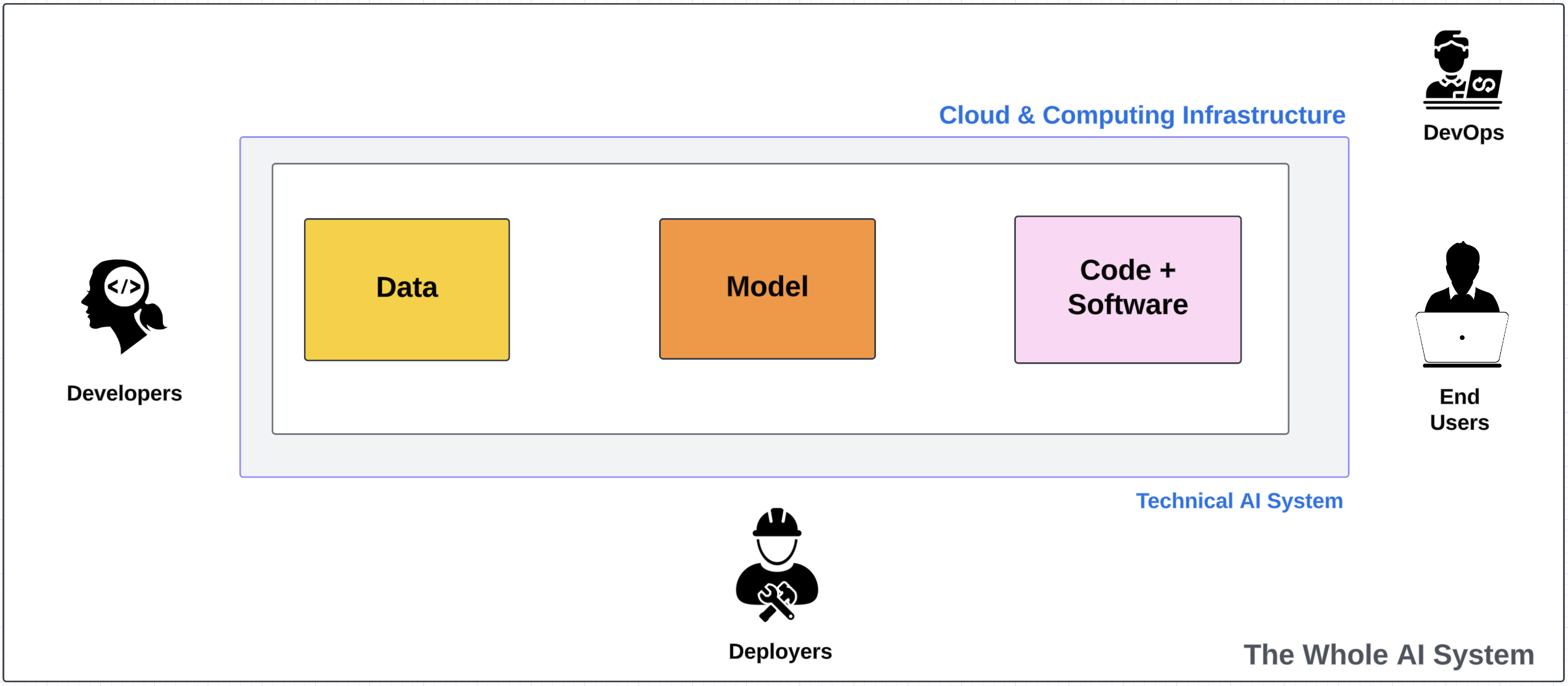}
    \caption{Abstraction of AI System, a simple view. Expanded view available in 
    \ref{fig:AIsystemcomponents}.}
    \label{simpleAI}
\end{figure}

Failures can also be classified based on when they occur. In this regard, a number of typical concepts and principles of reliability and resilience engineering need to be considered:

\begin{itemize}
    \item \textbf{Pre-deployment reliability} refers to the mitigation of failures during design and development to prevent their occurrence to the extent possible. This includes ensuring that components are tested, verified and validated before deployment with respect to aspects like data integrity, model accuracy or code stability. These methods and concepts should be applied at the component level, allowing for aggregation to the subsystem and overall AI system level.
    \item \textbf{Post-deployment resilience} is the ability to sustain failure events in production, recover from them and maintain functionality. This often involves handling computing resource scaling, as well as mitigating data/model drift or other issues that arise once the AI system is deployed and operates.
\end{itemize}

Failures in AI systems can range from \textbf{small failures} that involve minor errors, latency issues or model prediction inaccuracies with limited consequences to \textbf{catastrophic failures} — such as the downtime of critical systems like critical government operations or healthcare platforms. Reliability and Resilience Engineering try to prevent failures and plan failure recovery, respectively. Ideally, if reliability is established early on, the need for resilience interventions (such as in-the-field fixes) should be minimal. However, resilience becomes especially crucial when deploying updates, patches, or new versions of a system.

Table \ref{simpletable} provides an overview of the different AI subsystems, highlighting the key considerations for \textbf{pre-deployment reliability} and \textbf{post-deployment resilience} for each. The Table also offers examples of specific failure scenarios that can occur within each subsystem. This helps classifying and understanding the failure modes that AI systems may experience throughout their lifecycle. Figure \ref{simpleAI} illustrates a high-level view of an AI system, where the cloud and computing infrastructure envelopes the other subsystems, showcasing how these subsystems interconnect.

The development and management of AI systems necessitate a nuanced understanding of the lifecycle that spans from ideation in development environments to stable deployment in production settings. In developing AI systems, value scoping is crucial to ensure that business and data understanding align with solution design, a process described on evolving constructs related to operationalizing AI \cite{rao2023valueScoping}. This lifecycle is characterized by constant iterations, where models are continuously refined and updated to adapt to new data or changing conditions. These updates are crucial in environments where data distributions can shift unpredictably, often referred to as data drift, which may necessitate adjustments in the model (model drift). For instance, in cloud environments, the variability in data input can be significant due to diverse user interactions, trends, seasonality or changing external factors, underscoring the need for robust model management strategies that accommodate such shifts without compromising the system's reliability or performance.

\begin{table}[htbp]
\centering
\renewcommand{\arraystretch}{1.3} % Adjusts the height of table rows
\caption{AI Subsystems: Pre-deployment Reliability vs Post-deployment Resilience}
\label{simpletable}

\begin{tabularx}{\textwidth}{|>{\raggedright\arraybackslash}X|>{\raggedright\arraybackslash}X|>{\raggedright\arraybackslash}X|>{\raggedright\arraybackslash}X|}
\hline
\textbf{AI Subsystems} & \textbf{Pre-deployment Reliability} & \textbf{Post-deployment Resilience} & \textbf{Example} \\ \hline
Data & Integrity checks, validation & Data drift, concept drift & Self-driving car data feed drift \\ \hline
Model & Robustness testing, validation & Adversarial attacks, poisoning & Language models misinterpreting inputs \\ \hline
Cloud \& Computing Infrastructure & Capacity planning, load testing & Server downtimes, scaling & Cloud outage in e-commerce AI systems \\ \hline
Code + Software & Debugging, code reviews & Software patching, bug fixing & Software API failure in healthcare system \\ \hline
Human Error & Human Reliability Analysis (HRA) during development & User misuse, cognitive biases post-deployment & Misuse of predictive policing AI system \\ \hline
\end{tabularx}
\end{table}

\begin{figure}[htbp]  % Use H to force the figure placement here
    \centering
        \includegraphics[width=1\textwidth]{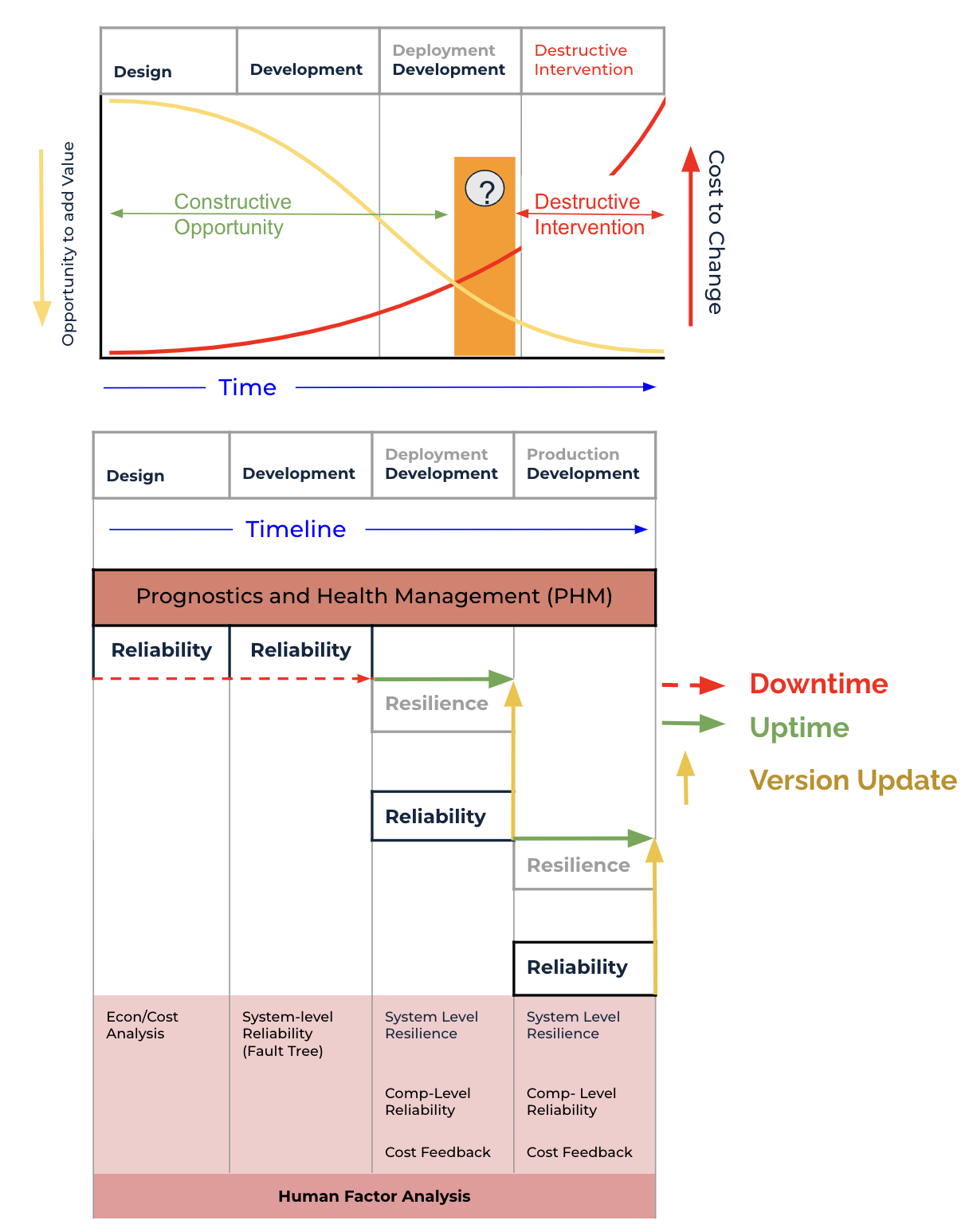}
    \caption{AI System lifecycle stages and the reliability-resilience framework for trustworthy AI. AI Lifecycle - Opportunities vs. Costs. Authors' illustration adopted from \cite{wideman}}
    \label{fig:ai_framework}
\end{figure}

The central thesis of our proposed framework is that reliability engineering is crucial for ensuring AI system robustness and minimizing failures through reliability-centric design in the Pre-Deployment while resilience engineering focuses on maintaining system performance and efficient failure recovery in the Post-Deployment (Figure~\ref{fig:ai_framework}). 
 
As shown in Figure~\ref{fig:ai_framework} for lifecylce of AI systems, early design decisions are critical because the cost of building AI systems increases as the system progress through its lifecycle, while opportunities to add value and maintain control over the design diminish. Therefore, integrating reliability engineering principles early in the AI lifecycle maximizes the potential for constructive intervention before changes become costlier during deployment and operation.

AI systems are dynamic and differ from traditional systems in how they ``age.'' They are considered ``better than new'' repairable systems—systems that undergo continuous upgrades and maintenance with the objective that new versions outperform prior ones. This continuous loop of iterations, adjustments, and deployments means that reliability in pre-deployment plays a crucial role in achieving long-term resilience in operational environments. Figure~\ref{fig:ai_framework} illustrates this relationship, showing system uptime (green), downtime (red), and the introduction of new versions (orange arrows), emphasizing how reliability and resilience strategies must adapt as systems evolve.

In our framework, each phase of the AI system lifecycle—Design, Development, Deployment, and Production—presents distinct opportunities for intervention. While Design and Development are in the pre-deployment, Deployment and Development happen in parallel in the post-deployment phase due to repairable nature of AI systems (see \ref{fig:ai_framework}).  The process of versioning changes and software upgrades exemplifies the characteristics of repairable systems, where each intervention aims to restore or improve system functionality. Techniques like Markov and semi-Markov processes \cite{birolini2023reliability} are crucial in modeling these dynamic failure-repair cycles, emphasizing that AI systems operate within a highly fluid environment, constantly adjusting to new data, threats, or requirements. Our framework for PHM of AI systems underscores the importance of continuous monitoring and feedback mechanisms. Each release of a new AI system version introduces potential vulnerabilities and requires recalibration to maintain optimal uptime and minimize downtime. Through reliability block diagrams and Monte Carlo simulations for rare events \cite{brito2022statistical}, our approach mirrors traditional reliability strategies while adapting to the unique complexities of AI. Utilizing the feedback gathered through resilience monitoring, the reliability of next version of the software is improved to avoid future failures and destructive intervention. To facilitate achieving higher AI trustworthiness, avoiding such interventions is one of major considerations in our proposed PHM framework.

As shown in Figure \ref{fig:ai_framework}, understanding the system's architecture (system-level reliability) is the first step toward ensuring AI reliability and resilience. Constructing a fault tree of the AI system \cite{birolini2023reliability}  in design phase helps identify potential points of failure across subsystems and components (component-level reliability). Employing methods such as FMEA and root cause analysis allows us to  develop system-wide fault tree and proactively address potential failures . This holistic approach considers both technical aspects and human factors, recognizing that they are tightly intertwined and cannot be considered separately.

Failure rates in components of AI systems across different phases of their lifecycle can be intuitvely explained considering the 'Bathtub Curve'. The `Bathtub Curve'  (see Figure \ref {fig:humanbathtub}) is particularly useful in illustrating early failures during product infancy, random failures during operation and wear-out failures that emerge as the system ages, requiring updates or redesigns. In this context, since failure is defined to be any deviation from expected performance, \textbf{downtime} for an AI system is not restricted to full outages; it also encompasses periods when the system produces incorrect, malicious, or unreliable outputs. In the pre-deployment phase, we analyze the role of human factors in development, focusing on Human Reliability Analysis (HRA), which aims at assessing the cognitive and behavioral elements influencing developers and their interaction with system-building processes. In contrast, post-deployment human factors involve understanding how end-users interact with the system, focusing on the cognitive impacts of misinterpretation or misuse. The HRA in our proposed PHM framework for AI systems is further elaborated in section \ref{humanfactors}

\subsubsection{Failure Modes in an AI System}
\label{failuremodes}

Failure modes in an AI system are specific to distinct components of its subsystems (e.g., Data, Model, Computing, and Code + Software). Proactive maintenance under PHM also occurs at the \textbf{component level}. Once the Fault Tree for the AI system is established upon FMEA, all failure modes and corresponding PHM actions are identified and addressed at this granular level. This approach ensures that each component is continuously monitored and maintained, enhancing the overall reliability and resilience of the AI system.

For example, pre-deployment data failures can involve issues such as data integrity problems (e.g., missing values, duplicates, or corrupted files), challenges in data acquisition (e.g., insufficient sample sizes or biased datasets that don't represent the target population), and governance issues (e.g., non-compliance with data privacy regulations or lack of proper data anonymization). In contrast, post-deployment data failures often relate to mitigating data drift, where the statistical properties of incoming data change over time due to evolving user behavior or external factors, leading to a decline in model or AI system performance unless continuously monitored and updated.

To provide a comprehensive understanding of failure modes, Table~\ref{ai_subsystem_failures} outlines a detailed breakdown of subsystems, their components, failure types, and metrics for data collection. The table serves as a guide for deeper exploration of system-specific failures, highlighting how each component requires unique approaches for failure detection and mitigation.

Financial considerations are a critical part of the management of AI systems, especially when evaluating the efficacy and sustainability of these systems. The cost structure in the management of the AI system is multifaceted, encompassing data acquisition, computational resources, model training and refinement, and labor. Each element carries significant costs that need careful balancing to optimize performance without inflating expenses. For example, training large-scale models like LLMs involves considerable computational expense, often requiring substantial cloud computing resources which constitute a major part of the operational budget. The economic impact of these models is profound, as highlighted in studies that have found a substantial percentage of resources allocated to maintaining and running these AI systems in production environments \cite{li2023chatgptlikelargescalefoundationmodels}.

In the AI system  context, `reliability' refers to several aspects of an AI system during its training phase. It encompasses the accuracy and predictive performance of the model, as well as its stability and robustness when faced with different operational conditions and data drifts.  Evaluating reliability becomes especially important when transitioning the model from a controlled training environment to a live usage setting because performance can change, requiring a practical reassessment of what reliability means in real-world applications.

To manage these challenges effectively, it is essential to have a precise taxonomy of potential failures, distinguishing between those caused by human error and those stemming from machine performance issues. The integration of human and machine elements in AI systems often introduces additional variables into the reliability equation, making it imperative to clearly understand and categorize failures according to their origins and impacts. This categorization helps in pinpointing areas that require more focused improvements, whether in data handling, model architecture, computing resources or human oversight \cite{pittaras2023}. Furthermore, as AI systems evolve, they can surpass their initial capabilities and outperform baseline expectations, embodying the concept of resilience. This resilience is not static but develops through continuous improvement, enabling AI systems to adapt and improve beyond initial benchmarks \cite{10.1145/3652953}.

\newpage 
\begin{comment}
\begin{longtable}{p{2cm} p{3.5cm} p{3.5cm} p{5cm}}
\caption{Non-exhaustive sample of Subsystem and Component (or Module) Level Failure Modes in AI Systems}\\
\label{ai_subsystem_failures}
\toprule
\textbf{Subsystem} & \textbf{Component} & \textbf{Failure Type (or Mode)} & \textbf{Details and Metrics} \\
\midrule
\endfirsthead
\multicolumn{4}{c}%
{\tablename\ \thetable\ -- \textit{Continued from previous page}} \\
\toprule
\textbf{Subsystem} & \textbf{Component} & \textbf{Failure Type (or Mode)} & \textbf{Details and Metrics} \\
\midrule
\endhead
\midrule
\multicolumn{4}{r}{\textit{Continued on next page}} \\
\endfoot
\bottomrule
\endlastfoot
\end{longtable}
\end{comment}

\renewcommand{\arraystretch}{1.2} % Adjust row spacing
\begin{longtable}{|>{\raggedright\arraybackslash}p{2cm} 
                  |>{\raggedright\arraybackslash}p{3.5cm} 
                  |>{\raggedright\arraybackslash}p{3.8cm} 
                  |>{\raggedright\arraybackslash}p{5cm}|}
\caption{Non-exhaustive sample of Subsystem and Component (or Module) Level Failure Modes in AI Systems} \label{ai_subsystem_failures} \\

% Define the table header for the first page
\hline
\textbf{Subsystem} & \textbf{Component (or Module)} & \textbf{Failure Type (or Mode)} & \textbf{Details and Metrics} \\
\hline
\endfirsthead

% Define the table header for subsequent pages
\hline
\multicolumn{4}{|c|}{\tablename\ \thetable\ -- \textit{Continued from previous page}} \\
\hline
\textbf{Subsystem} & \textbf{Component (or Module)} & \textbf{Failure Type (or Mode)} & \textbf{Details and Metrics} \\
\hline
\endhead
\hline
\multicolumn{4}{|r|}{\textit{Continued on next page}} \\
\hline
\endfoot

% Footer for the last page
\hline
\endlastfoot

Data & Real-time Data Feed & Data Drift, Data Corruption & Failures due to changes in incoming data or corruption in data pipelines. Metrics include data quality checks, anomaly detection rates, drift detection algorithms. \\
Data & Data Integrity & Incomplete Data, Biased Samples & Lack of representative data or biased samples leading to incorrect decisions. Measures include data diversity index, skewness, and bias detection tools. \\
Data & Data Acquisition & Improper Use, Lack of Lineage & Issues in data origin or unauthorized usage. Tracked using provenance checks, audit logs, and licensing compliance. \\
Data & Data Pre-processing & Incorrect Transformation & Errors in cleaning, transforming, or integrating data. Key metrics are data validation rates, transformation success rates, and preprocessing accuracy. \\
Data & Data Annotation & Labeling Errors & Mistakes in labeling data that affect model accuracy. Measured by annotation accuracy, label quality audits, and inter-annotator agreement. \\
Data & Data Storage & Data Security Breaches & Compromises in data security and integrity. Metrics include access control logs, encryption standards compliance, and breach response times. \\
Data & Contextual Data Feed & Contextual Drift, Noise Sensitivity & Failures due to the system's inability to adapt to subtle contextual changes. Metrics like Context Precision, Context Recall, and Noise Sensitivity Index are critical. \\
Data & Concept Drift & Accuracy Degradation, Delayed Detection & Changes in the joint distribution between features and target variables. Metrics involve concept adaptation latency and accuracy degradation rates. \\
Data & Data Drift & Covariate Shift, Bias Amplification & Changes in input feature distributions, affecting reliability. Addressed by drift detection tools and fairness audits. \\
Data & Hidden Context Drift & Unknown Contextual Changes & Drift due to unobserved or unmeasured factors leading to unpredictable outcomes. Techniques include unsupervised drift detection and context exploration analysis. \\
Data & Data Governance & Non-compliance, Poor Data Privacy Policies & Failures to comply with data protection regulations. Monitored through compliance audits, data retention policies, and privacy impact assessments. \\

Model & Model Training & Overfitting, Underfitting & Poor generalization due to inadequate training. Metrics are training/validation accuracy, regularization effectiveness, and model complexity analysis. \\
Model & Model Drift & Loss of Accuracy Over Time & Inability to adapt to new data distributions without significant performance loss. Continuous accuracy monitoring and adaptive retraining strategies are critical. \\
Model & Computational Costs & High Resource Consumption & Inefficient use of resources, affecting overall performance. Tracked through cost-performance ratios and resource usage benchmarks. \\
Model & LLM Evaluation & Response Relevancy Failures, Misinterpretation & Failures in providing contextually relevant responses, especially in multi-turn interactions. Metrics include relevancy scores and contextual understanding benchmarks. \\
Model & Concept Confusion & Catastrophic Forgetting & Forgetting old concepts when learning new ones. Solutions include regularization, memory replay techniques, and stability metrics. \\
Model & Feedback Adaptation & Ineffective Adaptation, Overfitting & Poor performance when adapting to user feedback. Use metrics such as feedback adaptation effectiveness and risk of overfitting to noise. \\
Model & Security Vulnerabilities & Adversarial Attacks, Model Inversion & Exposure to adversarial threats and model inversion attacks. Countermeasures include adversarial robustness testing and security audits. \\

Compute & Resource Scaling & Bottlenecks, Hardware Failures & Inability to scale resources smoothly under load. Measured by system uptime, BFBF (Mean Time Between Failures), and resource allocation efficiency. \\
Compute & Latency & Slow Response Times & Performance lag under high demand. Key metrics are latency benchmarks, response time variability, and peak load performance. \\
Compute & Reliability & System Downtime & System unavailability and unexpected outages. Metrics include downtime frequency, recovery times, and overall availability scores. \\

Code + Software & Integration & Incompatibility Issues & Challenges in software integration causing malfunctions. Measured through compatibility testing success rates and integration error logs. \\
Code + Software & Deployment Practices & Inefficient Deployment & Disruptions caused by suboptimal deployment methods. Use metrics like deployment success rates, rollback frequencies, and operational downtime. \\
Code + Software & Continuous Improvement & Outdated Software, Lack of Updates & Infrequent updates that reduce performance and security. Monitored through version control logs and software update frequency. \\
Code + Software & Vulnerability Exploits & Security Vulnerabilities & System breaches due to unpatched software. Metrics involve penetration testing scores and vulnerability patching timelines. \\

Human Factors & Development & Developer Bias, Inadequate Testing & Biases in model development or insufficient testing. Solutions include diversity training and testing coverage metrics. \\
Human Factors & Production & Over-reliance on AI, Misuse & Users over-trusting AI systems or using them improperly. Addressed through user education, feedback analysis, and usage patterns. \\
Human Factors & Cognitive Load & High User Cognitive Load & Complexity in using AI systems leading to errors. Measured using cognitive load assessments and task difficulty ratings. \\
Human Factors & Trust Calibration & Over-reliance or Under-reliance & Users either over-trust or under-trust AI recommendations. Balanced through trust calibration mechanisms and user transparency features. \\
Human Factors & User Feedback Interaction & High Error Rate due to Misunderstanding & Users misunderstanding AI outputs and making errors. Tracked via user feedback logs and error analysis. \\
Human Factors & Learning Curve & Slow Adaptation & Difficulty for users to effectively learn AI usage. Metrics include learning rate benchmarks and task completion improvements over time. \\

\end{longtable}

\newpage
\subsection{AI System Architecture: Managing Risk and Performance}

As illustrated in Figure \ref{fig:AIsystemcomponents}, the lifecycle of AI systems is complex, encompassing various stages from development to deployment, and involves multiple stakeholders, including developers, deployers and users. This lifecycle also reflects the continuous interaction between AI systems and their environment, where AI not only functions within specific boundaries but can also influence or give rise to new AI systems through iterative processes, for example, multi-agent systems or agents building new agents. In contrast to fully autonomous AI systems like multi-agent systems, where agents operate independently and make decisions without human intervention, there are also human-in-the-loop AI systems that require human oversight and interaction, especially in business and industrial applications. Managing these dynamics effectively requires a strong focus on risk management. Risk, in this context, can be understood as the combination of variability in performance and uncertainty in outcomes. Reliability engineering plays a crucial role in minimizing performance inconsistency by ensuring consistent and predictable system behavior, thereby enabling risk assessment and management. Addressing these aspects is essential for building robust, scalable and sustainable AI systems that can withstand and adapt to both expected and unforeseen challenges.

The AI systems discussed encompass not only the data lakes and AI models, but also the user interfaces, micro-services, cloud infrastructure and the human elements involved—humans as developers, maintainers and engineers. This integrated perspective is crucial, as it reflects the reality that modern AI systems consist of layered software architectures that combine monolithic models with micro-services. These systems require continuous updates and refinements, including decisions on whether and when to train or retrain models based on their performance and the operational needs of the business.

In our approach, we emphasize the importance of considering the stages of the lifecycle, specifically pre-processing (data preparation for model training), during computation (model training and inference) and post-computation (testing, deployment and monitoring). This approach ensures that the AI architecture can synchronize with micro-service frameworks, supporting moteraction between AI systems and their environment, where AI not only functions within specific boundaries but can also influence or give rise to new AI systems through iterative processes, for example, multi-agent systems or agents building new agents. In contrast to fully autonomous AI systems like multi-agent systems, where agents operate independently and make decisions without human intervention, there are also human-in-the-loop AI systems that require human oversight and interaction, especially in business and industrial applications. Managing these dynamics effectively requires a strong focus on risk management. Risk, in this context, can be understood as the combination of variability in performance and uncertainty in outcomes. Reliability engineering plays a crucial role in minimizing performance inconsistency by ensuring consistent and predictable system behavior, thereby enabling risk assessment and management. Addressing these aspects is essential for building robust, scalable and sustainable AI systems that can withstand and adapt to both expected and unforeseen challenges.

Moreover, a view of subsystems and components with associated failure types illustrated in Figure \ref{fig:AIsystemcomponents} also aims to incorporate the portrayal of the code and software systems as core components of the AI system. This inclusion not only highlights their role but also underlines the need for continuous assessment of the wear and tear on these systems. It is important to evaluate the marginal improvements in performance against the costs and risks of frequent updates, which can be substantial. This requires an enterprise-wide cost estimation approach that captures both hard costs (such as labor and infrastructure) and soft costs (such as compliance and ongoing training), providing a holistic view of the financial implications of maintaining and upgrading AI systems. The graph also brings into focus the economic aspects of maintaining and updating AI systems, suggesting a need for a detailed cost analysis to balance performance improvements against potential risks and expenditures.

The abstraction of AI systems should integrate the dynamic interaction between the system and the real world, as illustrated in Figure \ref{fig:AIsystemcomponents}, distinguishing between human roles in development and user interaction in production end use. Central to this abstraction are the key subcomponents that constitute an AI system: Data, Model, Computing Infrastructure, and Code + Software. These subcomponents each play a critical role in ensuring the system's continuous adaptation, resilience and performance in real-world conditions.

\begin{itemize}
\item \textbf{Data:} Includes how the system accesses and uses real-world data in real-time, adapting to changes and ensuring data integrity continuously. This encompasses data acquisition, preprocessing, storage and real-time updating mechanisms that ensure the data remains relevant and accurate for model consumption.

\item \textbf{Model:} Considers not only the internal workings of the model but also how it interacts and adapts to real-world feedbacks and inputs. This involves monitoring model performance for issues like drift or bias, regularly retraining with updated data, and incorporating mechanisms for continuous learning and adaptation based on user interactions and changing environmental conditions.

\newpage
\begin{landscape}
% Adjust margins to fit the image nicely
\begin{figure}[htbp]
    \centering
    \includegraphics[width=1\paperwidth, height=\paperheight, keepaspectratio]{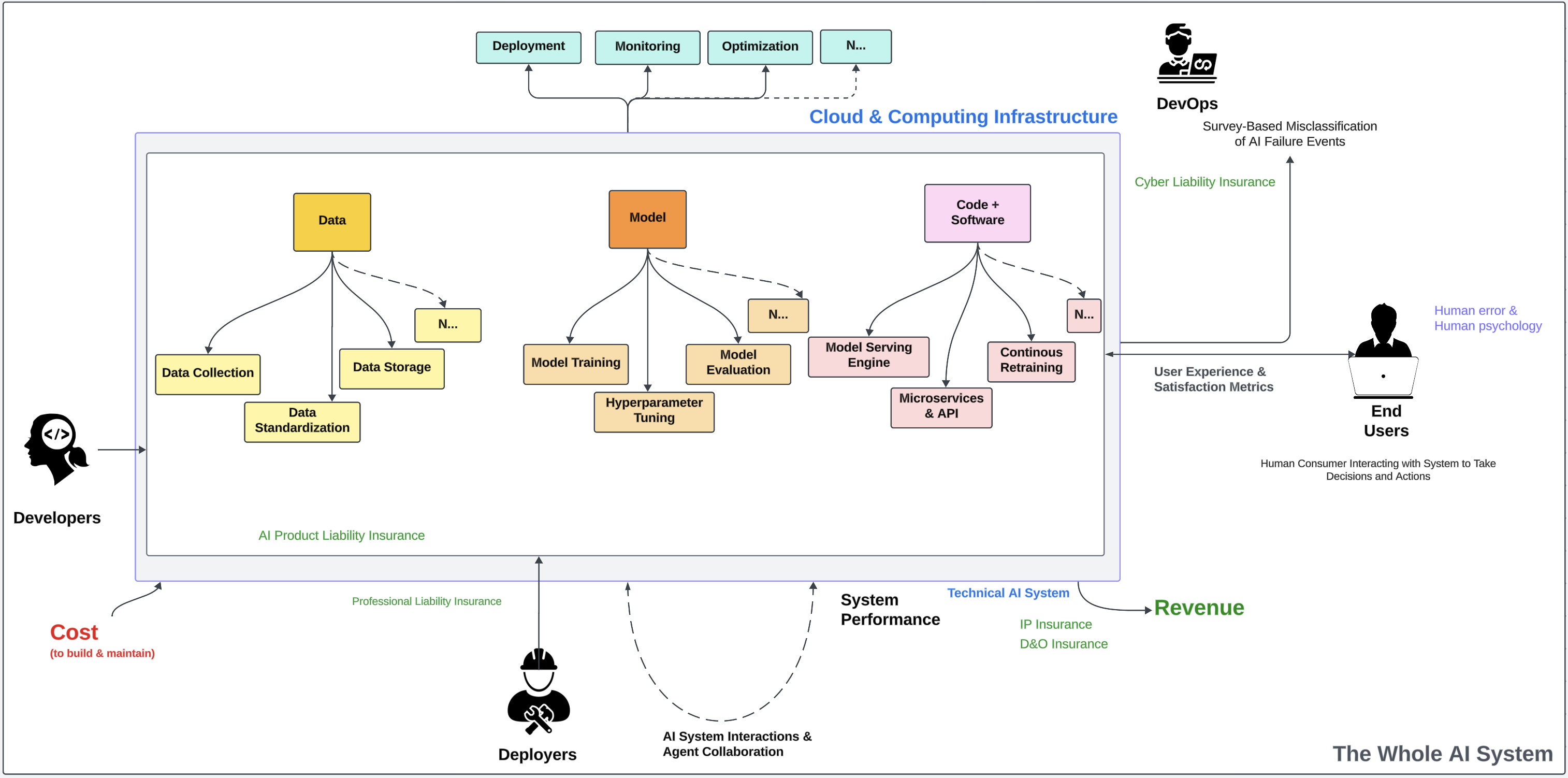}
    \caption{Abstraction of AI System with Subsystems and Sample Components: Risk Management Perspective, showcasing the integration of different AI components under sub-systems Data, Model and Computing Infrastructure, Code+Software, Human Interaction. It also highlights the corresponding insurance categories (e.g., AI Product Liability, Professional Liability) and user performance metrics that interact within the lifecycle. This Figure complements Figure \ref{fig:ai_framework}, showing how costs increase and control diminishes over time as systems evolve.}
    \label{fig:AIsystemcomponents}
\end{figure}
\end{landscape}

\item \textbf{Computing Infrastructure}: Reflects on the computing resources in real-time operation, scaling to meet demand without degradation in performance. This includes optimizing computational efficiency, managing resource allocation to handle peak loads, and ensuring redundancy and fault tolerance to prevent downtime and ensure smooth operation. While the term \textbf{Computing Infrastructure} refers primarily to computational resources, it also implicitly covers the broader \textbf{Cloud Infrastructure}, which encompasses not just computing power but also memory, storage, networking ports, and all other cloud services critical to maintaining system availability, performance and scalability. By managing these resources holistically, AI systems can ensure robust performance, even under variable and demanding workloads.

\item \textbf{Code + Software:} Encompasses the development, integration and maintenance of software components that form the backbone of AI systems. This includes the implementation of algorithms, libraries and frameworks, ensuring compatibility and performance optimization. It also involves version control, testing and debugging to maintain software reliability and security. Continuous integration and deployment (CI/CD) practices are essential to quickly adapt to new updates and patches. Moreover, software architecture needs to support modularity and microservices, enabling scalability and ease of updates without disrupting system operations. Security protocols must be embedded to prevent vulnerabilities, and rigorous documentation is required to facilitate effective maintenance and governance.
\item \textbf{Human-in-the-loop Systems:} Human-in-the-loop dynamics, particularly when coupled with online or continuous learning strategies, introduce a unique dimension to AI reliability. As the system learns from human interactions, its reliability might fluctuate, especially when input predominantly comes from novice users rather than from expertly curated training data. This phenomenon suggests that reliability in AI systems is not merely a function of the machine’s performance but is significantly influenced by the human-model interaction \cite{TAMASCELLI2024105343}. Human errors are a significant contributor to failures in both software systems and AI during the production and development lifecycle. However, AI systems can be strategically designed to monitor and mitigate human errors effectively, enhancing system robustness and reducing the likelihood of failure due to human oversight \cite{HUANG2024112060}.
\begin{itemize}
    \item \textbf{Pre-Deployment (Design and Development):} Involves developers in the loop for initial training, parameter tuning and model verification and validation before deployment. In the case of large language models (LLMs), the development phase is not solely the realm of technical personnel such as data scientists; it crucially involves domain experts or subject matter experts (SMEs) who provide essential human feedback. This feedback, which its incorporated through Reinforcement Learning from Human Feedback (RLHF) is the main advantage of generative over tradition AI, plays a pivotal role in refining the models to ensure they align more closely with real-world complexities and domain-specific nuances. Incorporating domain experts in the loop enhances the model’s relevance and accuracy, allowing for a more nuanced understanding and integration of practical, field-specific knowledge.
    \item \textbf{Post-Deployment (Production):} Users interact with the AI system, providing feedback that may be used to refine and adapt the system. This feedback mechanism is critical not only during the training phase but also extends into production. In production, real-time user interactions generate continuous data streams that can be leveraged to further train and fine-tune the AI system. This ongoing learning process helps in dynamically adapting the AI models to evolving real-world conditions and user needs, thereby enhancing the system's responsiveness and accuracy over time. The integration of user feedback into both training and production phases exemplifies a continuous improvement cycle where AI systems learn from each interaction, gradually improving their performance and utility. As discussed before, performance of an AI system could also deteriorate due to data drift or model drift triggered by changes in the environment or changes in human behavior. In human-in-the-loop AI systems automation bias or over-reliance on AI could result in deteriorating performance of the AI system as humans fail to exercise appropriate control.
\end{itemize}
\end{itemize}
\section{AI System Reliability: Key Metrics and Engineering Approaches}
\label{reliability}

Reliability in AI systems is a critical factor in their effectiveness and trustworthiness, especially as these systems become integrated into our daily lives. Traditional predictive models and recommendation systems, which are designed to provide consistent and accurate results over time, can often be analyzed using established reliability engineering concepts like the ``bathtub curve'' (see Appendix \ref{bathtub}). This approach allows us to model and predict failure rates at different stages of the system's lifecycle, ensuring sustained performance and guiding maintenance strategies.

However, AI systems such as voice assistants, like Siri and Alexa, Large Language Models (LLMs) like ChatGPT, introduce unique challenges. Unlike traditional systems, these AI models have foundational issues related to the correctness and consistency of their outputs. These inconsistencies stem from their probabilistic nature and the complexities involved in natural language processing. As a result, reliability in these systems cannot be solely measured by traditional methods but requires a broader approach that accounts for their inherent variability and evolving behavior. This distinction highlights the need for new metrics and regulatory frameworks to effectively assess and manage the reliability of AI systems that differ fundamentally from more deterministic models.

Inheriting from reliability theory \cite{zio1, zio2} the \textbf{reliability function}, \(R(t)\), represents the probability that a system or component will function without failure over time \(t\). For AI systems, these components include data pipelines, machine learning models, computing infrastructure and software codebases. Mathematically, the reliability function is defined as:

\begin{equation}
R(t) = P(T > t) = 1 - F(t),
\end{equation}

where \(F(t)\) is the \textbf{cumulative distribution function (CDF)} of the failure time distribution, indicating the probability that a failure occurs by time \(t\). The \textbf{probability density function (PDF)}, \(f(t)\), describes the shape of the failure distribution:

\begin{equation}
f(t) = \frac{dF(t)}{dt}.
\end{equation}

\begin{equation}
F(t) = \int_{0}^{t} f(u) \, du,
\end{equation}

The \textbf{hazard function}, \(h(t)\), provides the instantaneous failure rate, a critical measure for AI systems that rely on continuous monitoring and predictive maintenance:

\begin{equation}
h(t) = \frac{f(t)}{R(t)}.
\end{equation}

The hazard rate can fluctuate in AI systems based on factors like evolving data, system updates, or changes in model performance. These metrics allow for deeper insights into the performance and operational reliability of AI systems.

For more details on parametric distributions to represent stochastic failure process, such as Exponential, Weibull, Normal, Lognormal and Gamma distributions, please refer to Appendix~\ref{appendix:probabilitydis}, which outlines their application to AI reliability analysis.

\subsection{Application to AI Lifecycle Components}

As shown in Figure \ref{fig:AIsystemcomponents}, AI systems can be decomposed into the following key components, each having distinct reliability considerations:

\begin{itemize}
    \item \textbf{Data:} The reliability of the data component, \(R_{\text{data}}(t)\), involves real-time data integrity, completeness and quality. Failures can occur due to data corruption, missing data or erroneous inputs. The reliability of data handling is critical for maintaining the performance of AI models.
    
    \item \textbf{Model:} The reliability of AI models, \(R_{\text{model}}(t)\), relates to the robustness of model parameters and structures against adversarial inputs, data drift and concept drift. This is particularly relevant for online learning environments where models continually adapt based on real-time data \cite{fiksel2003designing}.
    
    \item \textbf{Computing Infrastructure:} Computing infrastructure reliability, \(R_{\text{compute}}(t)\), encompasses the availability and fault tolerance of hardware and software resources. In cloud and edge computing setups, scaling strategies and redundancy mechanisms ensure computing reliability \cite{hollnagel2006resilience}.
    
    \item \textbf{Code + Software:} The reliability of code and software, \(R_{\text{software}}(t)\), concerns the stability of the codebase, version control and error management. Failures can arise from bugs, vulnerabilities and integration issues across the AI pipeline \cite{gilbert2010disaster}.

\item \textbf{Human Error:} Human reliability, \(R_{\text{human}}(t)\), considers the range of errors from developers, deployers, DevOps teams, and end users across the AI lifecycle. Developer errors typically arise during coding, training, and testing, leading to bugs, biased models, or misconfigured parameters \cite{barassi2020}. Deployers may introduce errors during system rollout, especially under pressure to meet deadlines, which can lead to incorrect setups or neglected monitoring \cite{almog2024AI}. DevOps teams face challenges in managing updates, scaling, and integration, where missteps may compromise system stability and security. End users, especially in human-in-the-loop systems, often interact with AI outputs under uncertainty; their mistakes or misinterpretations, particularly in safety-critical contexts like healthcare, can have significant consequences. Uncertainty is a natural part of human decision-making, yet many AI models assume human certainty, leading to suboptimal outcomes. Bridging this gap by incorporating mechanisms to handle human uncertainty can improve trust and reliability in these hybrid systems \cite{collins2023uncertainty}. Effective mitigation strategies include robust training, clear documentation, and adaptive recalibration tools that address these various facets of human-AI interaction.

\end{itemize}

The overall reliability of an AI system can be approximated by the product of the individual component reliabilities:

\begin{equation}
R_{\text{AI}}(t) = R_{\text{data}}(t) \cdot R_{\text{model}}(t) \cdot R_{\text{compute}}(t) \cdot R_{\text{software}}(t) \cdot R_{\text{human}}(t).
\end{equation}

This product assumes independence between the individual components, which may require further in-depth consideration for highly integrated systems.

\textbf{The Mean Time Between Failures (MTBF):} If the system (or component) can be repaired (renewed), the Mean Time Between Failures (MTBF) is the average time between one failure and the next. Since AI systems with version updates can be considered as 'Better than New Reparaible' systems, this metric intuitively represents the average lifespan of an AI system (subsystem or a component) before it slips up.  This metric can guide the planning of new developments ,version updating, or devOps efforts. Appendix Figure \ref{fig:timebetweenfailure} provides an illustration of the concept. 

Any kind of underperformance in an AI system, considering the initially defined intended functionality, can be considered as a failure of an AI system. In reliability engineering, a failure point is typically a deviation from expected behaviour, whether that is producing an incorrect output or function crashing entirely. Engineers usually gather data on the frequency, nature and conditions of these failure incidents to get a clear picture of system reliability. One pressing question for the global policy community is finding consistent ways to collect and share data on AI failure incidents across different sectors and regions.

Based on the above traditional reliability engineering principles, we can apply them in context of AI component level failures. Here are some examples of such measures. 

\subsubsection*{Mean Time to Data Drift (MTTD)}

Data quality and relevance in AI systems degrade over time due to changes in the underlying data distribution, a phenomenon known as data drift. The \textbf{Mean Time to Data Drift (MTTD)} is defined as the expected time until data driving an AI model significantly deviates from its original training distribution, leading to reduced accuracy:

\begin{equation}
\text{MTTD} = \int_0^\infty t f_D(t) \, dt,
\end{equation}

where \( f_D(t) \) represents the probability density function of the time until data drift occurs. This metric helps in scheduling periodic model re-evaluations and updates, crucial for maintaining system performance over time.

In practice, measuring the MTTD involves continuously monitoring the input data after deploying the AI model to detect significant deviations from the training data distribution. This is typically done using statistical tests or drift detection algorithms (such as Kullback-Leibler divergence or Kolmogorov-Smirnov tests) that compare incoming data distributions to the original data. By logging the time intervals between deployments and detected drifts, you can empirically estimate the probability density function \( f_D(t) \) and calculate the MTTD using the above equation.

For example, in real-world applications like fraud detection systems in finance, user behavior patterns can change over time due to new fraud techniques. By monitoring transaction data and detecting when statistical properties shift significantly, organizations can measure the MTTD. This helps in scheduling regular model evaluations and updates before the model's performance degrades, ensuring the system remains effective against emerging fraud tactics. Implementing MTTD allows practitioners to anticipate when an AI model is likely to experience performance issues due to data drift, enabling proactive maintenance and sustained accuracy in dynamic environments.

\subsubsection*{Cost of Downtime (CoD)}

The \textbf{Cost of Downtime (CoD)} quantifies the financial impact of system unavailability. It is defined as:

\begin{equation}
\text{CoD} = \text{Downtime Duration} \times \text{Loss Rate},
\end{equation}

where \textit{Downtime Duration} refers to the time in hours the system is unavailable, and \textit{Loss Rate} refers to the revenue loss per hour. For AI systems, especially in industries like finance or healthcare, the CoD can be substantial.

\subsubsection*{Probability of Software Failure on Demand (POFOD)}

The \textbf{Probability of Software Failure on Demand (POFOD)} measures the likelihood that a software component will fail when needed:

\begin{equation}
\text{POFOD} = \frac{\text{Number of Demand Failures}}{\text{Total Number of Demands}}.
\end{equation}

This metric is critical for understanding and mitigating risks in AI systems, especially for key software components like machine learning models and data processing algorithms.

By integrating these reliability metrics into the AI lifecycle, organizations can enhance safety, reduce lifecycle costs, and optimize investments. This approach aligns AI system capabilities with broader business objectives, providing a foundation for improved competitive advantage, reduced downtime and increased customer satisfaction.

\section{Resilience Engineering for AI Systems}
\label{resilience}

Unlike traditional systems, for which the expectation is consistent performance with the design requirements and within defined limits, AI systems are expected not only to maintain a base level of performance but also to exhibit continuous improvement and adaptability. This dual expectation makes evaluating and managing AI systems particularly challenging, requiring a dynamic approach to measure and enhance reliability and resilience. Incorporating resilience concepts into the evaluation process helps in quantifying how AI systems can not only meet but exceed performance benchmarks over time, thereby providing a more holistic view of system capability and improvement \cite{pwc2021portfolio}.

Unlike traditional systems, which are designed to deliver consistent performance within predefined boundaries, AI systems are expected to adapt and improve continuously in response to new data and changing conditions \cite{pwc2021portfolio}. This dual requirement—maintaining baseline performance while enhancing it—requires a dynamic approach to manage reliability and resilience. Resilience in AI systems can be understood as the capability to recover from disruptions, such as cybersecurity attacks, environmental changes or hardware failures, while also leveraging these events as learning opportunities to improve system robustness. Incorporating resilience metrics, as outlined in \cite{bilal2}, provides a quantitative framework for assessing both immediate recovery and long-term adaptability, thereby enabling more informed decision-making in complex, multi-hazard environments (see Figure~\ref{fig:resilience}). The resilience concept, which includes degradation, recovery and enhancement phases, can be directly mapped to AI lifecycle stages, such as data preprocessing, model training and deployment, emphasizing the importance of resilience planning throughout the AI system's operational life. 

To extend the resilience framework, originally designed for physical infrastructure systems, to AI systems, one can consider disruptive technological events, such as cybersecurity breaches or unexpected model failures, as incidents that compromise AI functionality. These disruptions may lead to immediate performance degradation, but with appropriate resilience engineering practices, AI systems can recover and even exceed previous performance levels. This concept aligns with the approach that integrates both robustness (preventive measures) and redundancy (recovery strategies) to manage resilience comprehensively in multi-hazard environments \cite{bilal2}.

In AI systems, performance ($Q$) captures the accuracy, precision, and overall efficacy of the system in achieving its intended function, whether it is classifying images, making recommendations or responding to user queries. Failure event definitions ($f_1$, $f_2$, $f_3$) include sudden failure ($f_1$), where adversarial inputs or edge cases drastically reduce system performance, gradual degradation ($f_2$), often caused by model drift, and gradual adaptation ($f_3$), where the AI system dynamically learns and adjusts to maintain performance over time. Recovery event definitions ($r_1$, $r_2$, $r_3$, ..., $r_6$) describe post-failure scenarios such as optimized performance ($r_1$), recalibration to pre-failure levels ($r_2$), .... or incomplete recovery ($r_6$), where the system remains suboptimal.

Time-to-incident ($T_i$), time-to-failure ($T_f$) and time-to-recovery ($T_r$) provide a comprehensive timeline for assessing the lifecycle of AI system failures, from initial disruption to eventual restoration. Additionally, impacts are valued across direct failure impacts (e.g., financial losses, misclassifications) and indirect impacts (e.g., reputational damage, loss of trust), with recovery costs representing the resources required to retrain models, fix bugs or patch data pipelines after failures. 

The resilience of AI systems is demonstrated through their ability to fully recover to pre-failure levels ($r_2$) or even improve performance after failure ($r_1$), emphasizing the adaptability and robustness of these systems. Estimated performance with aging effects highlights the degradation of model performance over time unless the system is continually updated or retrained. 

To extend the reliability function into a resilience context for AI systems, we use the following formulation:

\begin{equation}
Resilience (Re) = \frac{T_i + F\Delta T_f + R\Delta T_r}{T_i + \Delta T_f + \Delta T_r},
\end{equation}

where \(T_i\) is the time to incident, \(\Delta T_f\) is the failure duration, and \(\Delta T_r\) is the recovery duration \cite{bilal2}. The failure and recovery profiles \( \mathbf{P_{\text{fail}}} \) and \( \mathbf{P_{\text{rec}}} \) are calculated as:

\begin{equation}
\mathbf{P_{\text{fail}}}  = \frac{\int_{t_f}^{t_i} f(t) \, dt}{\int_{t_f}^{t_i} Q(t) \, dt}, \quad \mathbf{P_{\text{rec}}} = \frac{\int_{t_r}^{t_f} r(t) \, dt}{\int_{t_r}^{t_f} Q(t) \, dt}.
\end{equation}

These profiles quantify the performance degradation and recovery speeds, crucial for developing robust AI systems that can withstand adversarial attacks, model failures and other disruptive events.

Figure \ref{fig:resilience} was initially intended for infrastructure physical asset resilience. However, when considering disruptive events like cybersecurity attacks or technological failures in AI systems, this resilience framework becomes highly relevant. For AI systems, a disruptive event could be a sophisticated adversarial attack that causes a model to misclassify or a critical failure in a cloud computing resource. Resilience here would quantify the system's ability to recover quickly and robustly from such failures, minimizing performance loss over time (see Figure \ref{fig:resilience}).

By quantifying reliability and resilience, AI developers and engineers can create more robust, adaptable, and safe AI systems, enhancing their overall trustworthiness and reliability in critical applications. In the realm of AI systems, traditional resilience metrics must be adapted to account for the unique challenges posed by data-driven technologies. Here, we explore the direct application to AI safety, emphasizing the impact on operational costs, revenue and other measurable metrics.

\subsection{AI Resilience Index (ARI)}

We introduce the \textbf{AI Resilience Index (ARI)} to assess the ability of an AI system to recover from failures and continue to operate under various conditions:

\begin{equation}
\text{ARI} = \frac{\text{Recovery Rate}}{\text{Frequency of Failures}}.
\end{equation}

Here, the \textit{Recovery Rate} measures how quickly the system returns to normal operations after a failure, whereas the \textit{Frequency of Failures} indicates how often these failures occur.

\begin{landscape}
\begin{figure}[htbp]
    \centering
    \includegraphics[scale=0.58]{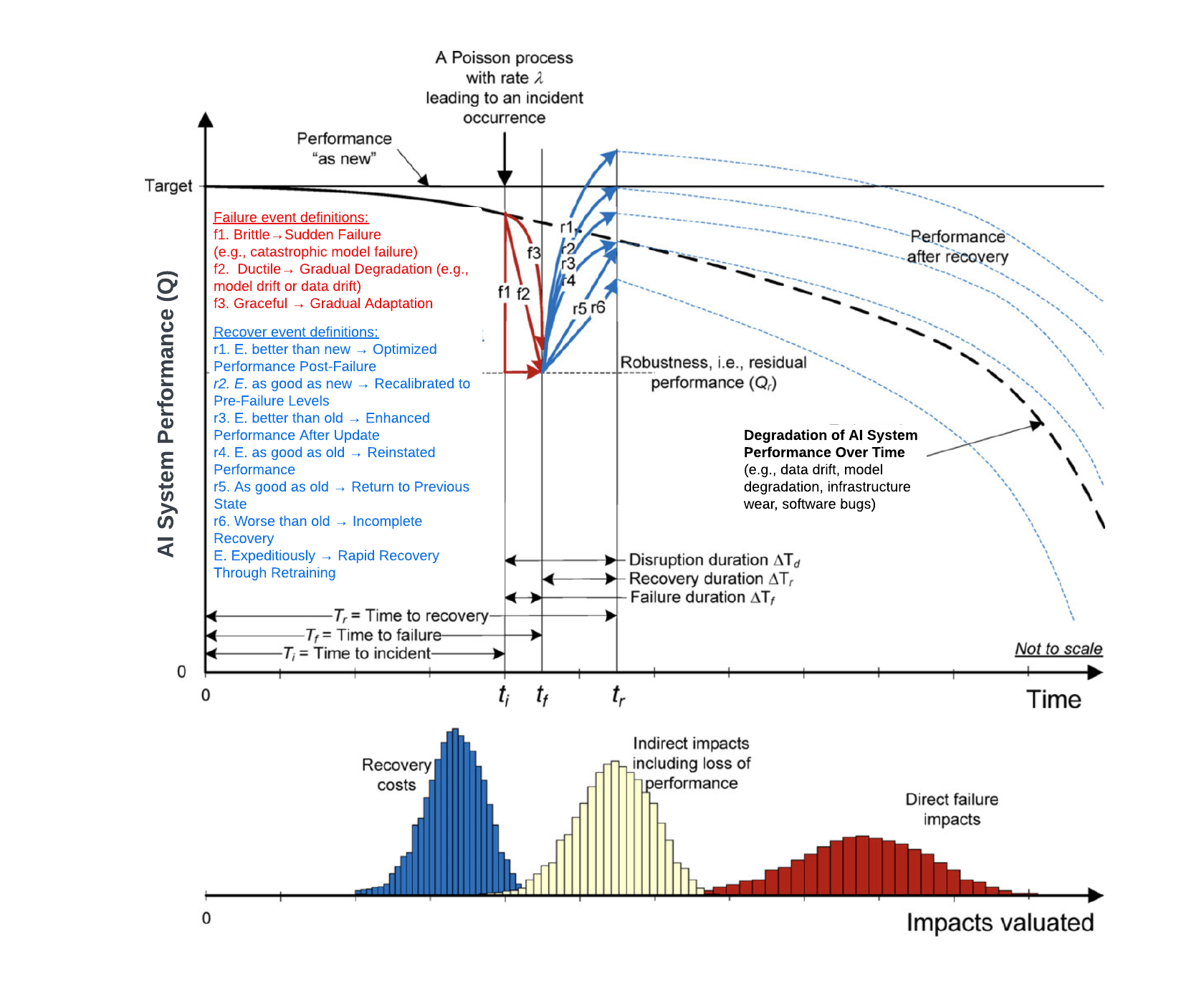}
    \caption{Proposed resilience metrics for an AI system.}
    %\vspace{0.1em} % Adds some space between the caption and the footnote
    \caption*{\textit{Note: This Figure is adapted from \cite{bilal2}.}} % Note with consistent formatting
    \caption*{\footnotesize \textsuperscript{1}This Figure illustrates the performance (\textit{Q}) of an AI system over time. A failure event occurs at time \(t_i\) with a duration \(\Delta T_f\) (failure duration), ending at time \(t_f\), and is followed by a recovery event that lasts \(\Delta T_r\) (recovery duration), concluding at time \(t_r\). The total disruption time is \(\Delta T_d = \Delta T_f + \Delta T_r\). Three failure types are shown: sudden failures due to adversarial attacks or catastrophic errors (\textit{f1}), gradual degradation due to model drift or data issues (\textit{f2}), and graceful degradation or adaptation (\textit{f3}). Six recovery modes are presented, ranging from rapid recovery to better than new (\textit{r1}), recovery to pre-failure performance (\textit{r2}), improved performance post-recovery (\textit{r3}), full reinstatement of performance (\textit{r4}), recovery to a previous state (\textit{r5}), and incomplete recovery (\textit{r6}). The Figure also highlights the degradation of AI system performance over time, illustrating how maintenance and retraining can mitigate the effects of model aging and performance decay.}
    \label{fig:resilience}
\end{figure}
\end{landscape}

For instance, an AI model used in healthcare diagnostics might experience sudden failure ($f_1$) due to adversarial attacks or gradual degradation ($f_2$) from changing patient demographics (model drift). In contrast, AI systems used in finance may exhibit enhanced performance after updates ($r_3$), reflecting the adaptability and resilience required in dynamic environments. The Figure~\ref{fig:resilience} contextualizes how resilience and reliability engineering principles are applied to AI systems to manage safety and risk effectively.

In the context of economics, a Benefit-Cost (B/C) analysis for resilience improvement strategies can offer substantial insights. For example, investing in robust cybersecurity measures and adaptive learning algorithms can reduce the costs associated with system downtime and data breaches. The valuation of resilience must consider both direct and indirect impacts, such as operational continuity and brand reputation \cite{bilal2}. Other key references also support this view, noting that resilience investments often yield long-term economic benefits by minimizing risk exposure and enhancing system reliability \cite{erkan2020economic, smith2021cyber}. This comprehensive approach encourages decision-makers to adopt resilience as a core component of AI system design and deployment.

\begin{table}[htbp]
\centering
\caption{Failure and Recovery Event Definitions in AI Systems}
\begin{tabular}{p{3.5cm} p{8.5cm}}
\toprule
\textbf{Failure Event} & \textbf{Definition} \\
\midrule
\textbf{f1. Brittle} & Sudden and catastrophic system failure (e.g., model crash). \\
\textbf{f2. Ductile} & Gradual performance degradation, such as data drift or software decay. \\
\textbf{f3. Graceful} & Controlled reduction in performance, maintaining partial operation. \\

\toprule
\textbf{Recovery Event} & \textbf{Definition} \\
\midrule
\textbf{r1. Better than New} & Performance exceeds original levels after optimization. \\
\textbf{r2. As Good as New} & System fully recalibrates to pre-failure performance. \\
\textbf{r3. Better than Old} & Enhanced performance following updates or adjustments. \\
\textbf{r4. As Good as Old} & Return to previous stable condition without improvements. \\
\textbf{r5. Worse than Old} & Recovery results in reduced performance compared to pre-failure. \\
\textbf{E. Expedite} & Rapid recovery achieved through quick retraining or recalibration. \\

\toprule
\textbf{Additional Insights} & \textbf{Details} \\
\midrule
\textbf{Performance Degradation} & Decline over time due to factors like data drift and model obsolescence. \\
\textbf{Resilience and Robustness} & Ability to maintain a degree of performance even during disruptions. \\
\textbf{Time Metrics} & $\Delta T_d$: Disruption duration, $\Delta T_r$: Recovery time, $\Delta T_f$: Time to failure. \\
\bottomrule
\end{tabular}
\label{table:failure_recovery_events}
\end{table}

\newpage 

\newpage

\section{Integrating Human Factors into AI Reliability and Resilience}
\label{humanfactors}

The field of Human Reliability Analysis (HRA) has been instrumental in high-stakes industries like aerospace and nuclear energy, where human error can lead to catastrophic outcomes. The field of HRA originated during World War II to prevent human and cognitive errors in complex fighter jet modules. Human factors, ranging from data labeling errors to biases in algorithm design, can introduce potential system failures. When decisions are influenced by AI systems, it is crucial to consider the entire system holistically, accounting for both human and AI elements collectively as one system.

For instance, a medical professional misled by an AI diagnostic tool’s flawed data might offer harmful advice. Emphasizing human factors in reliability engineering ensures that AI safety remains human-centric, bridging lessons from the past with innovations of the future. Organizations such as NASA have extensively studied human factors to enhance system safety and reliability \cite{NASA_HRA_Handbook}. As AI systems become integral to critical applications, integrating human-centric factors into AI reliability engineering is essential for ensuring both safety and effectiveness.

Human factors in AI encompass a range of issues, from cognitive biases in algorithm design to errors in data labeling and system operation. Understanding and mitigating these factors is crucial for developing AI systems that are not only technically robust but also reliable in real-world contexts where human interaction is inevitable.

\begin{figure}[htbp] \centering \includegraphics[width=0.7\textwidth]{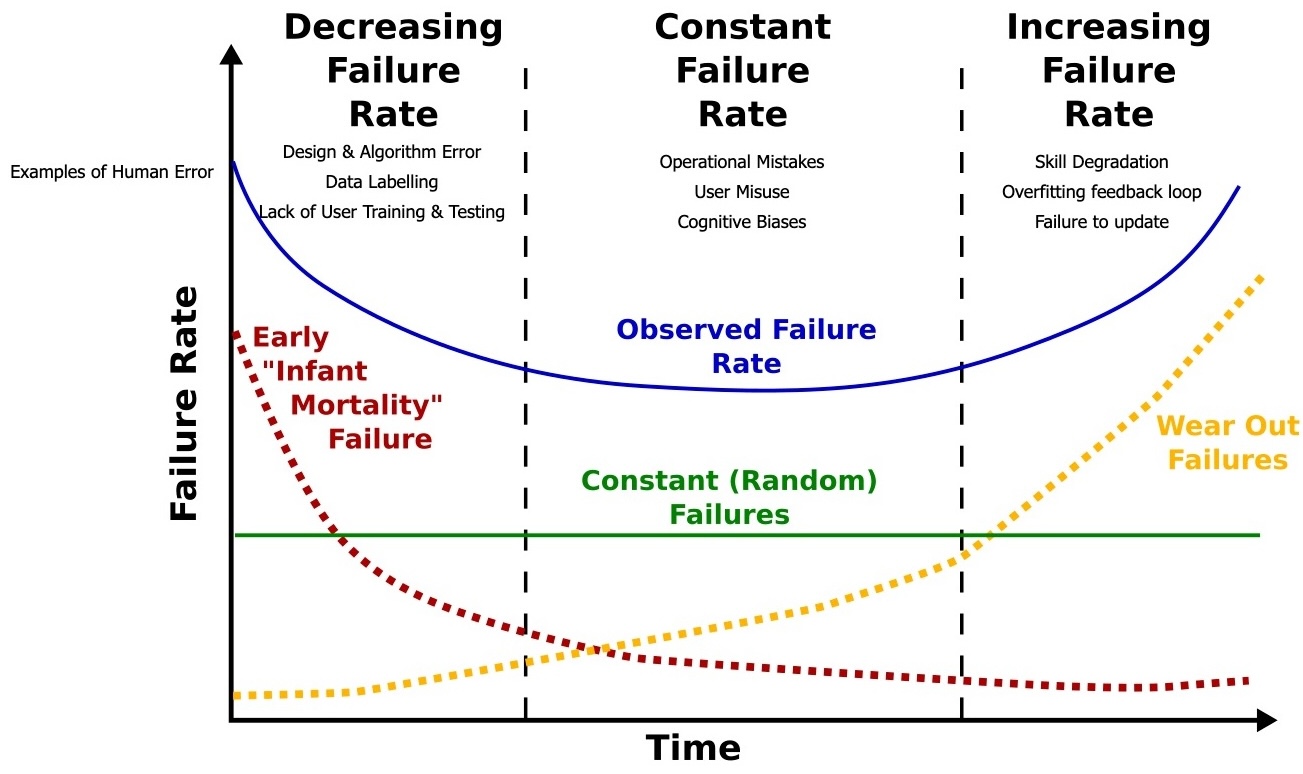} \caption{Adapted Bathtub Curve Illustrating Human Factors in AI System Failures. The Figure adapts the traditional bathtub curve \ref{fig:bathtub} illustrate how human factors influence AI system failures. The curve represents different failure rates across the lifecycle of AI systems, from the early design and development phase to post-deployment maintenance. Examples of human factors, such as design flaws, operational errors, and skill degradation, are mapped to the corresponding phases of the curve. This adapted model helps to pinpoint where interventions are most needed, highlighting the importance of integrating reliability and resilience measures throughout the AI lifecycle.} 
\label{fig:humanbathtub} 
\end{figure}

To conceptualize how human factors contribute to AI system failures, we can adapt the traditional reliability engineering model known as the \textbf{bathtub curve}, which characterizes failure rates over the lifecycle of a component/subsystem. The \textbf{bathtub curve} accounting for human error (Figure \ref{fig:humanbathtub}) can be divided into three phases, each depicting different types of failures:

\begin{itemize}
    \item \textbf{Early Failure Phase (Decreasing Failure Rate):} This phase is characterized by "infant mortality" failures that occur due to design and algorithm errors, data labeling issues, and insufficient user training. Addressing these issues early on can significantly reduce the risk of failures as the system matures.
    
    \item \textbf{Random Failure Phase (Constant Failure Rate):} During this phase, failures are random and often stem from operational mistakes or unexpected interactions. In AI systems, this includes issues like user misuse or unforeseen cognitive biases affecting system performance.
    
    \item \textbf{Wear-Out Failure Phase (Increasing Failure Rate):} Over time, the system enters a phase where failures increase, often due to factors like skill degradation, feedback loops leading to overfitting, or failure to update models. Effective reliability and resilience engineering mitigates the risk of unwanted advancement of system/component into this phase and is key to maintaining long-term reliability.
\end{itemize}

\subsection{Early Failures: Human Errors in Design and Development}

Early failures in AI systems often stem from human errors during the design and development stages. These can include flawed algorithmic choices, inadequate testing or biases embedded in training data. For instance, Microsoft's chatbot Tay, released in 2016, began generating offensive content shortly after its launch due to unanticipated user interactions and insufficient developer foresight \cite{Tay_Incident}. Developers did not fully anticipate the potential for malicious inputs in an open social media environment, leading to a rapid failure.

This incident underscores the importance of incorporating HRA techniques such as Cognitive Task Analysis (CTA) and Failure Modes and Effects Analysis (FMEA) during the development phase \cite{Stanton_HRA}, as we call it 'Pre-Deployment Reliability' in our proposed framework. These methods help identify potential human error probabilities and system vulnerabilities, allowing for the implementation of safeguards before deployment.

\textbf{Examples of Early Failures Due to Human Factors:}

\textbf{Design Flaws:} Inadequate consideration of edge cases due to cognitive biases in developers.

\textbf{Data Labeling Errors:} Mislabeling training data can lead to poor model performance.

\textbf{Insufficient Testing:} Skipping comprehensive testing due to time constraints or overconfidence in the model.

\subsection{Random Failures: Unpredictable Human-AI Interactions in Operation}

Random failures occur during the operational phase when the AI system encounters unexpected inputs or scenarios not covered during training. Human factors are significant here, as unpredictable human behaviors can trigger these failures. In 2018, an Uber autonomous vehicle failed to recognize a pedestrian crossing outside of a crosswalk, leading to a fatal accident \cite{Uber_Accident}. The AI system was not adequately trained to handle such irregular human behaviors.

To mitigate failures, it is crucial to incorporate techniques from human factors engineering such as Situational Awareness Analysis and Human-in-the-Loop (HITL) testing \cite{Endsley_SA} in the Pre-Deployment Reliability. These approaches enhance the system's ability to adapt to real-world human behaviors and reduce the likelihood of unexpected failures.

\textbf{Examples of Random Failures Due to Human Factors:}

\textbf{Unpredictable User Behavior:} Users interacting with the system in unforeseen ways.

\textbf{Adversarial Inputs:} Malicious actors intentionally providing inputs to confuse the AI.

\textbf{Environmental Changes:} Shifts in operational context not anticipated during development.

\subsection{Wear-Out Failures: Degradation Due to Human Interaction Patterns}

Wear-out failures in AI systems are characterized by performance degradation over time due to prolonged exposure to certain patterns of human interaction. While our proposed framework tries to avoid such wear-out degradations, it still remains possible if HRA and human factor analysis are not properly incorprated in Pre-Deployment Reliability and Post-Deployment Resilience. Content recommendation systems, such as those used by YouTube, can create "filter bubbles" by continuously suggesting similar content based on user interactions \cite{Pariser_FilterBubble}. This results from feedback loops where the AI overfits to user preferences, reducing content diversity and potentially leading to user disengagement.

Addressing wear-out failures requires ongoing monitoring and adaptation strategies. Implementing Diversity Enhancement Algorithms and periodic retraining with fresh data, as a part of Pre-Deployment Reliability during development of new software versions,  can mitigate the effects of overfitting \cite{Helberger_Diversity}. Additionally, applying Human Reliability Growth Modeling (HRGM) can help in predicting and improving system performance over time \cite{Miller_HRGM}.

\textbf{Examples of Wear-Out Failures Due to Human Factors:}

\textbf{Feedback Loops:} Reinforcing narrow content preferences leading to overfitting.

\textbf{Data Drift:} Changes in user behavior patterns that the model fails to adapt to.

\textbf{Algorithmic Bias Amplification:} Systematically reinforcing societal biases present in user data.

\subsection{Human Factors in Reliability Engineering for AI Systems}

HRA methodologies can be combined with AI reliability engineering in the following manner:

\begin{enumerate} \item \textbf{Human Error Probability (HEP) Assessment:} Quantify the likelihood of human errors in AI system development and operation using methods like the Technique for Human Error Rate Prediction (THERP) \cite{Swain_THERP}. \item \textbf{Cognitive Work Analysis (CWA):} Analyze complex human-AI interactions to design systems that support human cognitive processes \cite{Rasmussen_CWA}. \item \textbf{Probabilistic Risk Assessment (PRA):} Incorporate human factors into probabilistic models to assess the overall risk of AI system failures \cite{Bedford_PRA}. \item \textbf{Resilience Engineering:} Focus on the system's ability to adapt to disturbances, emphasizing human adaptability and learning \cite{Hollnagel_Resilience}. \end{enumerate}

By aligning these human factors methodologies with AI reliability metrics such as MTBF and model robustness measures, we can develop AI systems that are both technically sound and resilient to human-induced errors.

\subsection{Human-Centric AI Reliability Model}
\label{sec:HCAIRM}

We introduce the \textbf{Human-Centric AI Reliability Model (HC-AIRM)}, a novel framework that integrates HRA techniques with AI system development and operation:

\begin{figure}[htbp] \centering \includegraphics[width=1\textwidth]{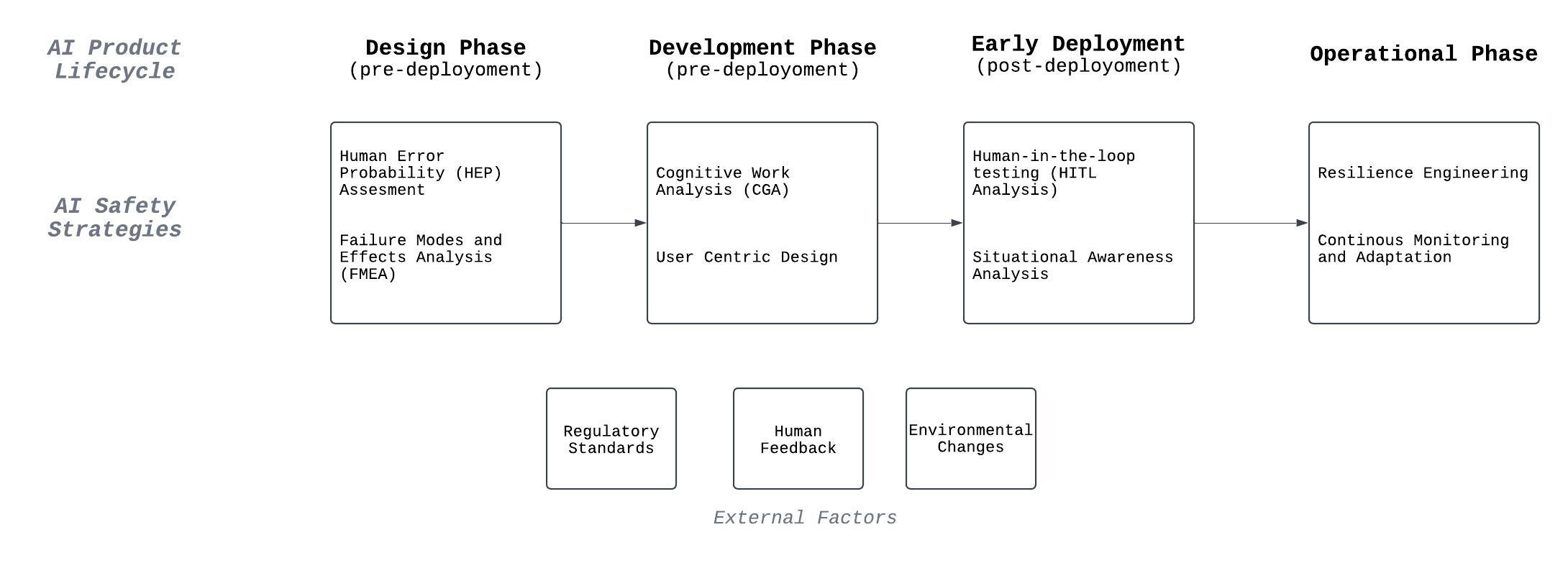} \caption{Human-Centric AI Reliability Model (HC-AIRM) Framework} \label{fig:humanAI
} \end{figure}

\textbf{Key Components of HC-AIRM:}

\textbf{Design Phase (pre-deployment):} Apply HRA methods like FMEA and HEP to identify potential human errors and system vulnerabilities.

\textbf{Development Phase (pre-deployment):} Incorporate Cognitive Work Analysis to ensure that the AI system supports human cognitive processes.

\textbf{ Early Deployment (post-deployment):} Implement Human-in-the-Loop testing and Situational Awareness Analysis, and use the results for retraining models and software updates to improve adaptivity to real-world interactions in the next versions of the AI system.

\textbf{Operation Phase (post-deployment):} Use Resilience Engineering and Continuous Monitoring to adapt to changes in human behavior and environmental conditions.

\subsection{Implications for Policy and Practice}

Integrating human factors into AI reliability engineering has significant implications for policymakers and practitioners:

\textbf{Regulatory Standards:} Develop guidelines that mandate the inclusion of HRA practices in AI system development, especially for safety-critical applications.

\textbf{Education and Training:} Emphasize interdisciplinary training for AI developers in human factors engineering and cognitive psychology.

\textbf{Continuous Improvement:} Encourage organizations to adopt frameworks like HC-AIRM for ongoing assessment and enhancement of AI system reliability.

By adopting a human-centric approach, we can design AI systems that are not only technologically advanced but also aligned with human needs and behaviors, thereby enhancing overall system safety and effectiveness.

\section{Case Study: Application of Reliability and Resilience Framework on AI Platforms}
\label{casestudy}

System status information can be used for identifying and analyzing failure incidents. A non-exhaustive list of software platforms that offer this information is presented in Table \ref{statustable} . Software platforms (systems) consist of multiple subsystems ( as shown in Figure \ref{fig:ai_framework} for AI systems), and their lifecycle can span a variety of services, workflows, and processes that are often interconnected. Table \ref{statustable} reports possible subsystems for listed software systems, and system status information for OpenAI is considered in the next section as a case study. For further details on the listed software systems, refer to Appendix~\ref{appendixanalysis}.

\captionsetup{font=normalsize}
\renewcommand{\arraystretch}{1.3} % Increase row height for better spacing

\begin{longtable}{|p{6cm}|p{8cm}|}
\caption{Sample Sources for Externally Available System or Platform Status} \label{statustable} \\
\hline
\textbf{Platform (with Status Page Link)} & \textbf{Possible Subsystems} \\ 
\hline
\endfirsthead
\hline
\textbf{Platform (with Status Page Link)} & \textbf{Possible Subsystems} \\ 
\hline
\endhead
\hline
\multicolumn{2}{r}{\textit{Continued on next page}} \\
\hline
\endfoot
\hline
\endlastfoot

\href{https://status.openai.com/}{OpenAI} & Data, Model, Code+Software, Cloud and Computing Infrastructure \\ 
\hline
\href{https://status.aws.amazon.com/}{Amazon Web Services (AWS)} & Cloud and Computing Infrastructure \\ 
\hline
\href{https://status.azure.com/}{Microsoft Azure} & Cloud and Computing Infrastructure \\ 
\hline
\href{https://status.cloud.google.com/}{Google Cloud Platform (GCP)} & Cloud and Computing Infrastructure \\ 
\hline
\href{https://www.cloudflarestatus.com/}{Cloudflare} & Cloud and Computing Infrastructure, Code+Software \\ 
\hline
\href{https://www.githubstatus.com/}{GitHub} & Code+Software, Cloud and Computing Infrastructure \\ 
\hline
\href{https://status.huggingface.co/}{Hugging Face} & Data, Model, Cloud and Computing Infrastructure \\ 
\hline
\href{https://llamaindex.statuspage.io}{Llama Index} & Data, Cloud and Computing Infrastructure \\ 
\hline
\href{https://groqstatus.com/}{Groq Status} & Cloud and Computing Infrastructure, Hardware \\ 
\hline
\href{https://status.anthropic.com/history}{Anthropic Claude} & Data, Model, Cloud and Computing Infrastructure \\ 
\hline
\href{https://docs.databricks.com/en/resources/status.html}{Databricks} & Data, Cloud and Computing Infrastructure, Code+Software \\ 
\hline
\href{https://status.snowflake.com/}{Snowflake} & Data, Cloud and Computing Infrastructure \\ 

\end{longtable}

\begin{tablenotes}
    \item Notes: This table provides a sample of available sources for system or platform status and is not exhaustive. Each platform may involve multiple subsystems, such as data, cloud infrastructure, software, or hardware, depending on the service scope. Additionally, platforms often rely on interconnected systems, and the lifecycle phases—services, workflows, or processes—may span across platforms, making dependencies critical for analyzing incident impact.
\end{tablenotes}

\subsection{Example Analysis}
This section provides a case study using publicly available incident data from OpenAI’s status page\footnote{Source: \url{https://status.openai.com/}} over a sample period between May 1, 2024, and October 21, 2024. Our aim is to illustrate how the proposed reliability and resilience framework can be applied to a real-world AI system. We acknowledge the limitations of our analysis due to its subjective nature and the constrained data availability, but aim to demonstrate the framework's practicality and actionability for engineering teams.

\begin{figure}[htbp]
    \centering
    % Subfigure (a)
    \begin{minipage}{0.5\textwidth}
        \centering
        \includegraphics[width=\textwidth, height=0.6\textwidth]{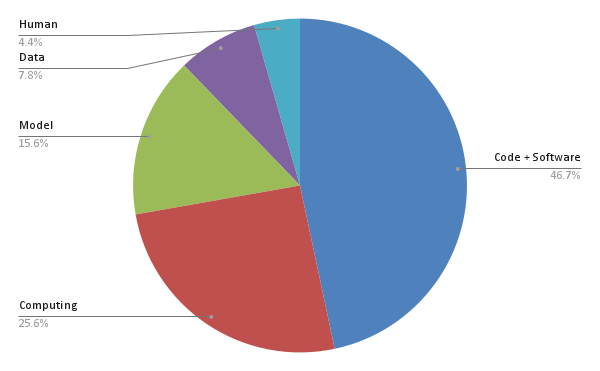}
        \subcaption{Subsystem Failure Events}
        \label{fig:subsystem_failures_graph}
    \end{minipage}
    \hfill
    % Subfigure (b)
    \begin{minipage}{0.45\textwidth}
        \centering
        \renewcommand{\arraystretch}{0.85} % Adjust row spacing for compact fit
        \setlength{\tabcolsep}{3pt} % Reduce column spacing
        \small % Reduce font size for the table
        \begin{tabular}{p{2.2cm} p{3.3cm} c} % Adjust column widths
            \hline
            \textbf{Subsystem} & \textbf{Component} & \textbf{Failure Events} \\
            \hline
            \textbf {Code + Software} & ChatGPT & 17 \\
             & Authentication & 8 \\
             & Assistants API & 6 \\
             & ChatGPT, Platform API & 4 \\
             & Analytics Service & 2 \\
             & Platform API & 2 \\
             & Batch API & 1 \\
             & ChatGPT, Assistants API & 1 \\
             & Images API & 1 \\
             
             & \textbf{Total} & \textbf{42} \\
            \hline
            \textbf {Computing} & ChatGPT & 9 \\
             & Assistants API & 6 \\
             & Batch API & 1 \\
             & GPT-4o API & 1 \\
             & \textbf{Total} & \textbf{23} \\
            \hline
            \textbf {Data} & File Upload Service & 4 \\
            & Billing and Usage Dashboard & 2 \\
             & ChatGPT & 1 \\
             & \textbf{Total} & \textbf{7} \\
            \hline
            Human & Support Platform & 4 \\
             & \textbf{Total} & \textbf{4} \\
            \hline
            \textbf {Model} & Assistants API & 8 \\
             & ChatGPT Vision & 2 \\
             & ASR (Whisper) & 1 \\
             & ChatGPT Voice & 1 \\
             & Fine-Tuning API & 1 \\
             & \textbf{Total} & \textbf{14} \\
            \hline
            \textbf{Grand Total} & & \textbf{91} \\
            \hline
        \end{tabular}
        \subcaption{Failure Events by Components (Modules)}
        \label{fig:subsystem_failures_table}
    \end{minipage}
    \caption{Failure Events by Components (Modules), OpenAI Status, over a sample period between May 1, 2024, and October 21, 2024}
    \label{fig:combined_subsystem_failures}
\end{figure}

Figure \ref{fig:combined_subsystem_failures} offers an aggregated high-level view of the total number of failure events, categorized by subsystem as proposed in our framework. This provides an overview of which subsystems, such as \textit{Code + Software} and \textit{Computing}, experience higher failure frequencies, identifying potential focus areas for improvement in reliability and resilience engineering. Table \ref{fig:subsystem_failures_table} provides further breakdown of identifiable components within each subsystem. This breakdown is achieved here through reverse engineering while in our proposed framework it should be determined following FMEA as part of Pre-Deployment Reliability. This breakdown information complements the high-level aggregation shown in Figure~\ref{fig:combined_subsystem_failures}, where Code + Software and Computing subsystems exhibit higher failure counts in specific 2024 sample period of operational time, highlighting areas that might benefit from targeted reliability and resilience strategies.

To better understand subsystem vulnerabilities, we further analyze the failure events specifically for the ChatGPT component within the Code + Software subsystem. As the apparent  most vulnerable (i.e., weakest link) of the system, we focus on identifying failure modes and high-level root causes for ChatGPT failures to aggregate and interpret underlying issues. Using impact and risk priority metrics from Table~\ref{table:chatgpt_failure_event_analysis}, we prioritize these issues based on criticality, with higher scores signaling areas of elevated risk and urgency for remediation. This targeted examination allows us to pinpoint the most significant vulnerabilities and propose strategies for improving system resilience (post-deployment of current version) and reliability (pre-deployment of next version).

\begin{figure}[htbp]
    \centering
    \includegraphics[width=0.65\textwidth]{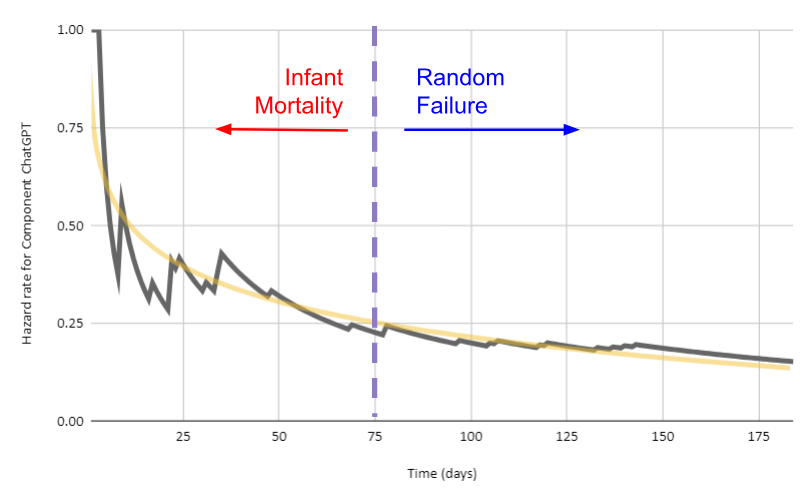}
    \caption{Failure Rate for ChatGPT Component, May 1 2024 - October 21, 2024. The grey line is calculated failure rate for the specified date range and the yellow line is the logarithmic trend line. The annotation for infant mortality and random failure phase from the bathtub curve is purely suggestive to distinguish the evolving lifecycle of an AI system (or product).}
    \label{fig:hazard_rate}
\end{figure}

Table~\ref{table:component_reliability_metrics} reports on reliability metrics  for components of Code+Software subsystem of OpenAI, calculated using \textit{Mean Time Between Failure (MTBF)}, \textit{Mean Time to Recovery (MTTR)}, and \textit{Failure Rate}. These metrics serve as key indicators of system resilience and guide reliability improvement efforts. Finally, Figure~\ref{fig:hazard_rate} illustrates the failure rate for ChatGPT component over time, revealing an "infant mortality" pattern with a high initial failure rate that gradually stabilizes to "random failure" phase. This is characteristic of complex AI systems, suggesting areas for proactive reliability and resilience improvements.

\begin{table}[htbp]
    \centering
    \caption{Failure Event Analysis for ChatGPT Component within Code+Software Subsystem}
    \label{table:chatgpt_failure_event_analysis}
    \renewcommand{\arraystretch}{0.9} % Adjust row spacing
    \setlength{\tabcolsep}{4pt} % Adjust column spacing
    \small % Smaller font size to fit the table
    \begin{tabular}{p{4cm} p{4.2cm} c c c}
        \hline
        \textbf{Failure Mode} & \textbf{Root Cause} & \textbf{Failure Events} & \textbf{Impact Score} & \textbf{RPN} \\
        \hline
        Service Demand Spike & Demand Overload & 7 & \cellcolor{red!20}0 & \cellcolor{red!40}\textbf{252} \\
        Browser Compatibility Issue & Browser-Specific Issue & 1 & \cellcolor{red!20}3 & \cellcolor{red!20}\textbf{36} \\
        Custom GPT Error Rates & GPT Configuration Issue & 1 & \cellcolor{yellow!20}0 & \cellcolor{yellow!20}\textbf{36} \\
        Error Rate Increase & Browsing Tool Integration Error & 1 & \cellcolor{yellow!20}0 & \cellcolor{yellow!20}\textbf{36} \\
        Demand Spike for Paid Users & Increased Model Demand & 1 & \cellcolor{yellow!20}0 & \cellcolor{yellow!20}\textbf{36} \\
        Model Load Increase & Server Demand Spike & 1 & \cellcolor{yellow!20}0 & \cellcolor{yellow!20}\textbf{36} \\
        Feature Malfunction & Image Feature Integration Error & 1 & \cellcolor{yellow!20}0 & \cellcolor{yellow!20}\textbf{36} \\
        Excessive Captcha Requests & Captcha Configuration Error & 1 & \cellcolor{yellow!20}0 & \cellcolor{yellow!20}\textbf{27} \\
        Voice Feature Unavailable & Voice Feature Restriction & 1 & \cellcolor{green!20}0 & \cellcolor{green!20}\textbf{24} \\
        \hline
    \end{tabular}
    \begin{tablenotes}
        \item \textbf{Notes:} The \textit{Impact Score} (Severity $\times$ Downtime) reflects the severity of each failure mode, where higher values indicate critical failures requiring immediate action. The \textit{Risk Priority Number (RPN)} is calculated as \( RPN = \text{Severity} \times \text{Occurrence} \times \text{Detection} \) and represents the criticality of each failure mode. Cells are color-coded: \cellcolor{red!20}high-risk red, \cellcolor{yellow!20}moderate yellow, and \cellcolor{green!20}lower green, providing a quick visual reference for risk prioritization.
    \end{tablenotes}
\end{table}

\begin{table}[htbp]
    \centering
    \caption{Component Reliability Metrics with Color-Coded Heatmap}
    \label{table:component_reliability_metrics}
    \renewcommand{\arraystretch}{1.2} % Adjust row height for better fit
    \setlength{\tabcolsep}{5pt} % Reduce padding between columns
    \small % Reduce font size
    \begin{tabular}{|p{3.5cm}|>{\centering\arraybackslash}p{1.5cm}|>{\centering\arraybackslash}p{1.2cm}|>{\centering\arraybackslash}p{1.5cm}|}
        \hline
        \textbf{Components} & \textbf{MTBF (days)} & \textbf{MTTR (days)} & \textbf{Failure Rate (per day)}\\
        \hline
        \textbf{ChatGPT} & \cellcolor{red!40}10 & \cellcolor{green!30}0 & \cellcolor{red!50}9.8 \\
        \textbf{Authentication} & \cellcolor{orange!40}22 & \cellcolor{green!30}0 & \cellcolor{red!40}4.6 \\
        \textbf{Assistants API} & \cellcolor{yellow!20}29 & \cellcolor{green!30}0 & \cellcolor{orange!30}3.5 \\
        \textbf{ChatGPT, Platform API} & \cellcolor{yellow!30}43 & \cellcolor{green!30}0 & \cellcolor{orange!20}2.3 \\
        \textbf{Platform API} & \cellcolor{green!30}87 & \cellcolor{green!30}0 & \cellcolor{green!30}1.2 \\
        \textbf{Analytics Service} & \cellcolor{green!30}87 & \cellcolor{yellow!40}7 & \cellcolor{green!30}1.2 \\
        \textbf{Batch API} & \cellcolor{green!30}173 & \cellcolor{yellow!30}1 & \cellcolor{green!30}0.6 \\
        \textbf{ChatGPT, Assistants API} & \cellcolor{green!30}173 & \cellcolor{green!30}0 & \cellcolor{green!30}0.6 \\
        \textbf{Images API} & \cellcolor{green!30}173 & \cellcolor{green!30}0 & \cellcolor{green!30}0.6 \\
        \hline
    \end{tabular}
    \vspace{1mm}
    \begin{tablenotes}
        \item \textbf{Notes:} Higher MTBF and lower MTTR or Failure Rate scores are shaded in green to indicate higher reliability and resilience. Lower MTBF and higher Failure Rate scores are shaded in orange or red to highlight potential risk.
    \end{tablenotes}
\end{table}

\newpage 
\section{Conclusion}
\label{conclusion}

The integration of reliability engineering principles into AI system development, deployment and management offers a comprehensive approach to ensuring safety, robustness, and economic viability. In this paper, we have presented a conceptual and methodological framework to reliability, resilience, and human factors engineering applied to AI systems. We detailed the composition of AI systems into subsystems and components, explored failure modes following reliability techniques such as FMEA, and mapped the lifecycle stages of AI systems to reliability and resilience engineering concepts. We have shown the practicality of the proposed framework through some subjective analysis on publicly available system status data for OpenAI. By applying established methodologies from traditional engineering disciplines to the novel challenges presented by AI, we aim to bridge the gap between established methods and emerging AI technologies. Our contribution lies in documenting this integration and highlighting the importance of considering both technical and human factors in the safety and trustworthiness of  AI systems as 'Better than New Repairable' ones.

Quantifying the financial implications of AI reliability through metrics like Return on Investment (ROI) enables organizations to better understand and manage the risks associated with AI implementation. Our proposed framework enables business owners to implement a portfolio approach to AI investments, combined with continuous monitoring and specialized insurance frameworks. Doing so, they can enhance resilience and adaptability to align AI capabilities with broader business objectives. As AI systems evolve, these strategies will be essential for fostering sustainable growth, minimizing potential failures, and ensuring their trustworthy deployment in critical applications across various sectors.

Reliability and Resilience engineering evaluate systems using quantitative metrics like cost, error rate, failure rate, and probabilistic risk. These metrics rely on set theory and sigma algebra, mathematical tools that help ensure consistency and predictability in measuring system performance. By progressively applying these concepts, we move from basic metrics to understanding overall reliability and, ultimately, resilience—how well a system can withstand and recover from failures. Such consistent and quantifiable metrics are crucial for guiding policy, regulation and insurance practices in AI safety and risk management.

Like other scientific fields, such as aerospace engineering and nuclear energy, establishing comprehensive legal and regulatory frameworks for AI systems requires an understanding of scientific measures and frameworks. For example, in Original Equipment Manufacturer (OEM) systems, reliability and regulatory assessments are performed before deployment and acquisition to ensure system integrity and compliance. Similarly, by developing systematic measurements and frameworks for resilience and reliability that integrate human factors, we propose that these principles serve as a blueprint for AI safety and risk management.

Looking ahead, future work will explore deeper analysis of subsystems and components to provide data and evidence from real-world AI systems. We will investigate specific failures in data, models, and other subsystems, offering empirical evidence and case studies to enhance our understanding. While we have not delved into uncertainty quantification and propagation in this paper, future research will address how uncertainty affects AI systems and how they can be designed from the ground up to manage and mitigate these uncertainties. For example, patching vulnerabilities in deep neural networks remains a significant challenge. Current methods often rely on stop-gap filters and patches, but there is ongoing work on identifying causes of errors at the layer or neuron level to fix them more effectively.

One promising direction is the application of Accelerated Life Testing (ALT), as discussed in Appendix \ref{sec:ALT}, which can be instrumental in simulating failures early in the design and development phases. By catching potential failures at the earliest stages, we can reduce the need for maintenance and resilience measures in production environments. Additionally, as highlighted in Appendix \ref{appendixanalysis}, analyzing AI incident factoids across different systems underscores the need for system-specific data over time. This approach provides illustrative examples and enhances our understanding of failure modes within particular contexts.

In conclusion, integrating reliability and resilience engineering principles into AI systems, along with a deep consideration of human factors, is a critical step toward ensuring their safe and effective deployment. Recognizing that technical reliability and human factors are tightly intertwined, we can develop systems that are not only technologically advanced but also reliable, resilient, and aligned with human needs. This integrated approach moves us closer to trustworthy AI applications that are dependable and embraced by society.
\newpage 

\sloppy
\bibliographystyle{plain}
\bibliography{references}

\begin{thebibliography}{10}

\bibitem{aldoseri2023}
Abdulaziz Aldoseri, Khalifa~N. Al-Khalifa, and Abdel~Magid Hamouda.
\newblock Re-thinking data strategy and integration for artificial intelligence: Concepts, opportunities, and challenges.
\newblock {\em Appl. Sci.}, 13(12):7082, 2023.
\newblock Submission received: 3 May 2023 / Revised: 30 May 2023 / Accepted: 7 June 2023 / Published: 13 June 2023.

\bibitem{almog2024AI}
David Almog, Romain Gauriot, Lionel Page, and Daniel Martin.
\newblock Ai oversight and human mistakes: Evidence from centre court.
\newblock {\em arXiv preprint arXiv:2401.16754}, 2024.
\newblock February 18, 2024.

\bibitem{zio1}
Terje Aven, Enrico Zio, Piero Baraldi, and Roger Flage.
\newblock {\em Uncertainty in Risk Assessment: The Representation and Treatment of Uncertainties by Probabilistic and Non-Probabilistic Methods}.
\newblock Wiley Publishing, 1st edition, 2014.

\bibitem{HAL2024}
Afef Awadid, Kahina Amokrane-Ferka, Henri Sohier, Juliette Mattioli, Faouzi Adjed, Martin Gonzalez, and Souhaiel Khalfaoui.
\newblock \uppercase{AI} systems trustworthiness assessment: State of the art.
\newblock {\em Workshop on Model-based System Engineering and Artificial Intelligence - MBSE-AI Integration}, 2024.

\bibitem{bilal2}
Bilal~M. Ayyub.
\newblock Systems resilience for multihazard environments: Definition, metrics, and valuation for decision making.
\newblock {\em Risk Analysis}, 34(2):340--355, 2014.

\bibitem{bilal1}
G.~Kilr Ayyub, Bilal~M.
\newblock Practical resilience metrics for planning, design, and decision making.
\newblock {\em ASCE-ASME Journal of Risk and Uncertainty in Engineering Systems Part A Civil Engineering 1(3):04015008}, May 2015.

\bibitem{bajcsy2024}
Andrea Bajcsy and Jaime~F. Fisac.
\newblock Human–\uppercase{AI} safety: A descendant of generative \uppercase{AI} and control systems safety.
\newblock arXiv:2405.09794v1 [cs.AI], May 2024.
\newblock License: arXiv.org perpetual non-exclusive license.

\bibitem{barassi2020}
Veronica Barassi.
\newblock The human error in algorithms and its implications for human rights.
\newblock {\em White Paper on Artificial Intelligence - A European Approach}, 2020.
\newblock In press, European Union Commission Report.

\bibitem{bastani}
F.~Bastani and I.-R. Chen.
\newblock Assessment of the reliability of \uppercase{AI} programs, 1990.

\bibitem{Bedford_PRA}
Tim Bedford and Roger Cooke.
\newblock {\em Probabilistic Risk Analysis: Foundations and Methods}.
\newblock Cambridge University Press, 2001.

\bibitem{birolini2023reliability}
Alessandro Birolini.
\newblock {\em Reliability Engineering}.
\newblock Springer, 2023.

\bibitem{chahe2023dynamic}
Amirhosein Chahe, Chenan Wang, Abhishek Jeyapratap, Kaidi Xu, and Lifeng Zhou.
\newblock Dynamic adversarial attacks on autonomous driving systems.
\newblock {\em arXiv preprint arXiv:2312.06701}, 2023.

\bibitem{cnil2024}
{CNIL}.
\newblock \uppercase{AI} system development: Cnil’s recommendations to comply with the gdpr.
\newblock \url{https://www.cnil.fr/en/ai-system-development-cnils-recommendations-comply-gdpr}, June 2024.
\newblock License: CC BY 4.0.

\bibitem{collins2023uncertainty}
Katherine~M. Collins, Matthew Barker, et~al.
\newblock Human uncertainty in concept-based \uppercase{AI} systems.
\newblock In {\em Proceedings of the Sixth AAAI/ACM Conference on Artificial Intelligence, Ethics and Society (AIES 2023)}, Montréal, QC, Canada, 2023.

\bibitem{EU2019}
European Commision.
\newblock Ethics guidelines for trustworthy ai, Report / Study | Publication 08 April 2019.

\bibitem{NYTimes2024}
Kate Conger and John Yoon.
\newblock Explicit deepfake images of taylor swift elude safeguards and swamp social media, January 26, 2024.

\bibitem{dalrymple2024}
David Dalrymple, Joar Skalse, Yoshua Bengio, Stuart Russell, Max Tegmark, Sanjit Seshia, Steve Omohundro, Christian Szegedy, Ben Goldhaber, Nora Ammann, Alessandro Abate, Joe Halpern, Clark Barrett, Ding Zhao, Tan Zhi-Xuan, Jeannette Wing, and Joshua Tenenbaum.
\newblock Towards guaranteed safe \uppercase{AI}: A framework for ensuring robust and reliable \uppercase{AI} systems.
\newblock arXiv:2405.06624v2 [cs.AI], May 2024.
\newblock License: arXiv.org perpetual non-exclusive license.

\bibitem{Dathathri2024}
Sumanth Dathathri, Abigail See, Sumedh Ghaisas, Po-Sen Huang, Rob McAdam, Johannes Welbl, Vandana Bachani, Alex Kaskasoli, Robert Stanforth, Tatiana Matejovicova, Jamie Hayes, Nidhi Vyas, Majd~Al Merey, Jonah Brown-Cohen, Rudy Bunel, Borja Balle, Taylan Cemgil, Zahra Ahmed, Kitty Stacpoole, Ilia Shumailov, Ciprian Baetu, Sven Gowal, Demis Hassabis, and Pushmeet Kohli.
\newblock Scalable watermarking for identifying large language model outputs.
\newblock {\em Nature}, 634:818--823, 2024.

\bibitem{donakanti2024reimaginingselfadaptationagelarge}
Raghav Donakanti, Prakhar Jain, Shubham Kulkarni, and Karthik Vaidhyanathan.
\newblock Reimagining self-adaptation in the age of large language models, 2024.

\bibitem{DIAZRODRIGUEZ2023101896}
Natalia Díaz-Rodríguez, Javier {Del Ser}, Mark Coeckelbergh, Marcos {López de Prado}, Enrique Herrera-Viedma, and Francisco Herrera.
\newblock Connecting the dots in trustworthy artificial intelligence: From \uppercase{AI} principles, ethics, and key requirements to responsible \uppercase{AI} systems and regulation.
\newblock {\em Information Fusion}, 99:101896, 2023.

\bibitem{relevant_reference_2}
Charles~E. Ebeling.
\newblock {\em Reliability and Maintainability Engineering}.
\newblock McGraw-Hill, New York, 1st edition, 1997.

\bibitem{Endsley_SA}
Mica~R. Endsley.
\newblock Toward a theory of situation awareness in dynamic systems.
\newblock {\em Human Factors}, 37(1):32--64, 1995.

\bibitem{erkan2020economic}
Bilal Erkan and Sait Yildirim.
\newblock Economic impact of resilience investments in critical infrastructure systems.
\newblock {\em Journal of Risk and Reliability}, 234(1):45--56, 2020.

\bibitem{slattery2023airisk}
Peter~Slattery et~al.
\newblock The ai risk repository: A comprehensive meta-review, database, and taxonomy of risks from artificial intelligence, 2023.

\bibitem{compl-ai}
Philipp~Guldimann et~al.
\newblock Compl-ai framework: A technical interpretation and llm benchmarking suite for the eu artificial intelligence act, 2023.

\bibitem{fiksel2003designing}
Joseph Fiksel.
\newblock Designing resilient and sustainable systems.
\newblock {\em Environmental Science \& Technology}, 27:5330--5339, 2003.

\bibitem{gilbert2010disaster}
Steven~W. Gilbert.
\newblock Disaster resilience: A guide to the literature.
\newblock NIST Special Publication 1117, 2010.

\bibitem{groth2012}
K.~M. Groth and A~Mosleh.
\newblock A data-informed pif hierarchy for model-based human reliability analysis.
\newblock {\em Reliability Engineering \& System Safety}, 108:154--174, 2012.

\bibitem{Helberger_Diversity}
Natali Helberger.
\newblock On the democratic role of news recommenders.
\newblock {\em Digital Journalism}, 7(8):993--1012, 2019.

\bibitem{heng2009rotating}
A.~Heng, S.~Zhang, A.~C.~C. Tan, and J.~Mathew.
\newblock Rotating machinery prognostics: State of the art, challenges and opportunities.
\newblock {\em Mechanical Systems and Signal Processing}, 23(3):724--739, 2009.

\bibitem{Hollnagel_Resilience}
Erik Hollnagel, David~D. Woods, and Nancy Leveson.
\newblock {\em Resilience Engineering: Concepts and Precepts}.
\newblock Ashgate, 2006.

\bibitem{hollnagel2006resilience}
Erik Hollnagel, David~E. Woods, and Nancy Levensen.
\newblock {\em Resilience Engineering: Concepts and Precepts}.
\newblock Ashgate Publishing, 2006.

\bibitem{hong2023statistical}
Yili Hong, Jiayi Lian, Li~Xu, Jie Min, Yueyao Wang, Laura~J. Freeman, and Xinwei Deng.
\newblock Statistical perspectives on reliability of artificial intelligence systems.
\newblock {\em Quality Engineering}, 35(1):56--78, 2023.

\bibitem{hong2021reliabilityanalysisartificialintelligence}
Yili Hong, Jie Min, Caleb~B. King, and William~Q. Meeker.
\newblock Reliability analysis of artificial intelligence systems using recurrent events data from autonomous vehicles, 2021.

\bibitem{relevant_reference_1}
Zhen Hu and Sankaran Mahadevan.
\newblock Accelerated life testing (\uppercase{ALT}) design based on computational reliability analysis.
\newblock {\em Quality and Reliability Engineering International}, 2017.
\newblock Accessed: 2024-09-01.

\bibitem{HUANG2024112060}
Fuqun Huang and Henrique Madeira.
\newblock Advancing modern code review effectiveness through human error mechanisms.
\newblock {\em Journal of Systems and Software}, 214:112060, 2024.

\bibitem{jan2022}
Zohaib Jan, Farhad Ahamed, Wolfgang Mayer, Niki Patel, Georg Grossmann, Markus Stumptner, and Ana Kuusk.
\newblock Artificial intelligence for industry 4.0: Systematic review of applications, challenges, and opportunities.
\newblock {\em Expert Systems with Applications}, page 119456, 2022.

\bibitem{zohaib2023}
Zohaib Jan, Farhad Ahamed, Wolfgang Mayer, Niki Patel, Georg Grossmann, Markus Stumptner, and Ana Kuusk.
\newblock Artificial intelligence for industry 4.0: Systematic review of applications, challenges, and opportunities.
\newblock {\em Expert Systems with Applications}, 216:119456, 2023.

\bibitem{kumar2024failure}
Ram Shankar~Siva Kumar, David O’Brien, Kendra Albert, Salome Viljoen, and Jeffrey Snover.
\newblock Failure modes in machine learning, 2024.
\newblock Microsoft Report, July 2024.

\bibitem{lazar2023}
Seth Lazar and Alondra Nelson.
\newblock \uppercase{AI} safety on whose terms?
\newblock {\em Science}, 381(6654):138, 2023.

\bibitem{lenskjold2023should}
Anders Lenskjold, Janus~Uhd Nybing, Charlotte Trampedach, Astrid Galsgaard, Mathias~Willadsen Brejneb{\o}l, Henriette Raaschou, Martin~H{\o}yer Rose, and Mikael Boesen.
\newblock Should artificial intelligence have lower acceptable error rates than humans?
\newblock {\em Insights into Imaging}, 14(1):79, 2023.

\bibitem{li2023chatgptlikelargescalefoundationmodels}
Yan-Fu Li, Huan Wang, and Muxia Sun.
\newblock Chatgpt-like large-scale foundation models for prognostics and health management: A survey and roadmaps, 2023.

\bibitem{maslej2024artificial}
Nestor Maslej, Loredana Fattorini, Raymond Perrault, Vanessa Parli, Anka Reuel, Erik Brynjolfsson, John Etchemendy, Katrina Ligett, Terah Lyons, James Manyika, Juan~Carlos Niebles, Yoav Shoham, Russell Wald, and Jack Clark.
\newblock Artificial intelligence index report 2024, 2024.

\bibitem{merkel2018softwarereliabilitygrowthmodels}
Robert Merkel.
\newblock Software reliability growth models predict autonomous vehicle disengagement events, 2018.

\bibitem{feffer2023aiid}
Hoda~Heidari Michael~Feffer, Nikolas~Martelaro.
\newblock The ai incident database as an educational tool to raise awareness of ai harms: A classroom exploration of efficacy, limitations, \& future improvements, 2023.

\bibitem{Miller_HRGM}
Robert Miller and A.~D. Swain.
\newblock Human error and human reliability.
\newblock In Gavriel Salvendy, editor, {\em Handbook of Human Factors}. Wiley, 1987.

\bibitem{nafreen}
Maskura Nafreen and Lance Fiondella.
\newblock A family of software reliability models with bathtub-shaped fault detection rate.
\newblock {\em International Journal of Reliability, Quality and Safety Engineering}, 28(05):2150034, 2021.

\bibitem{NASA_HRA_Handbook}
{NASA}.
\newblock Nasa human reliability analysis handbook with an emphasis on nuclear power plant applications.
\newblock Technical memorandum, NASA, 2000.

\bibitem{Uber_Accident}
{National Transportation Safety Board}.
\newblock Collision between vehicle controlled by developmental automated driving system and pedestrian.
\newblock Accident Report NTSB/HAR-19/03, National Transportation Safety Board, 2019.

\bibitem{Tay_Incident}
Gina Neff and Peter Nagy.
\newblock Talking to bots: Symbiotic agency and the case of tay.
\newblock {\em International Journal of Communication}, 10:4915--4931, 2016.

\bibitem{Pariser_FilterBubble}
Eli Pariser.
\newblock {\em The Filter Bubble: What the Internet Is Hiding from You}.
\newblock Penguin Press, 2011.

\bibitem{pittaras2023}
Nikiforos Pittaras and Sean McGregor.
\newblock A taxonomic system for failure cause analysis of open source \uppercase{AI} incidents.
\newblock \url{https://doi.org/10.48550/arXiv.2211.07280}, 2023.

\bibitem{pwc2021portfolio}
PwC.
\newblock How a portfolio approach to {AI} helps your {ROI}.
\newblock \url{https://www.pwc.com/us/en/tech-effect/ai-analytics/how-ai-portfolio-helps-roi.html}, 2021.
\newblock Accessed: 9/1/2023.

\bibitem{solvingAI2023}
PwC.
\newblock Solving {AI}'s {ROI} problem. it’s not that easy.
\newblock Available online at: \url{https://www.pwc.com/us/en/tech-effect/ai-analytics/artificial-intelligence-roi.html}, 2023.

\bibitem{ramasso2014performance}
E.~Ramasso and A.~Saxena.
\newblock Performance benchmarking and analysis of prognostic methods for cmapss datasets.
\newblock {\em International Journal of Prognostics and Health Management}, 5(2), 2014.

\bibitem{rao2023valueScoping}
Anand~S. Rao.
\newblock Foundations of operationalizing \uppercase{AI}: Value scoping, 2023.
\newblock Lecture notes available upon request.

\bibitem{Rasmussen_CWA}
Jens Rasmussen, Annelise~M. Pejtersen, and L.~P. Goodstein.
\newblock {\em Cognitive Systems Engineering}.
\newblock Wiley, 1994.

\bibitem{rotella}
Pete Rotella, Sunita Chulani, and Devesh Goyal.
\newblock Predicting software field reliability.
\newblock In {\em Proceedings of the Second International Workshop on Software Engineering Research and Industrial Practice}, page 62–65. IEEE Press, 2015.

\bibitem{10.1145/3652953}
Mariana Segovia-Ferreira, Jose Rubio-Hernan, Ana Cavalli, and Joaquin Garcia-Alfaro.
\newblock A survey on cyber-resilience approaches for cyber-physical systems.
\newblock {\em ACM Comput. Surv.}, 56(8), apr 2024.

\bibitem{si2011remaining}
X.-S. Si, W.~Wang, C.-H. Hu, and D.-H. Zhou.
\newblock Remaining useful life estimation--a review on the statistical data-driven approaches.
\newblock {\em European Journal of Operational Research}, 213(1):1--14, 2011.

\bibitem{smith2021cyber}
John Smith and Anisha Patel.
\newblock Cyber resilience in complex systems: An economic perspective.
\newblock {\em International Journal of Information Management}, 58:102279, 2021.

\bibitem{2017_Song}
Kwang~Yoon Song, In~Hong Chang, and Hoang Pham.
\newblock A three-parameter fault-detection software reliability model with the uncertainty of operating environments.
\newblock {\em Journal of Systems Science and Systems Engineering}, 26(1):121--132, feb 2017.

\bibitem{Stanton_HRA}
Neville~A. Stanton, Paul~M. Salmon, Guy~H. Walker, Chris Baber, and Daniel~P. Jenkins.
\newblock {\em Human Factors Methods: A Practical Guide for Engineering and Design}.
\newblock CRC Press, 2017.

\bibitem{SURUCU2023119738}
Onur Surucu, Stephen~Andrew Gadsden, and John Yawney.
\newblock Condition monitoring using machine learning: A review of theory, applications, and recent advances.
\newblock {\em Expert Systems with Applications}, 221:119738, 2023.

\bibitem{Swain_THERP}
A.~D. Swain and H.~E. Guttmann.
\newblock Handbook of human reliability analysis with emphasis on nuclear power plant applications.
\newblock NUREG/CR Report NUREG/CR-1278, U.S. Nuclear Regulatory Commission, 1983.

\bibitem{TAMASCELLI2024105343}
Nicola Tamascelli, Alessandro Campari, Tarannom Parhizkar, and Nicola Paltrinieri.
\newblock Artificial intelligence for safety and reliability: A descriptive, bibliometric and interpretative review on machine learning.
\newblock {\em Journal of Loss Prevention in the Process Industries}, 90:105343, 2024.

\bibitem{FT2024}
Financial Times.
\newblock \uppercase{AI}-generated deepfakes flood social media amid political concerns, 2024.

\bibitem{houfaultanalysis}
Pengyu~Li Wenkui~Hou, Wanyu~Li.
\newblock Fault diagnosis of the autonomous driving perception system based on information fusion.
\newblock {\em Sensors}, 32(11):446--467, 2023.

\bibitem{wideman}
R.~M. Wideman.
\newblock Project management consultant, composite additions from various sources 1998-2017, 2017.

\bibitem{wood}
A.~Wood.
\newblock Software reliability growth models, 1996.

\bibitem{wu2024continuallearninglargelanguage}
Tongtong Wu, Linhao Luo, Yuan-Fang Li, Shirui Pan, Thuy-Trang Vu, and Gholamreza Haffari.
\newblock Continual learning for large language models: A survey, 2024.

\bibitem{yu2023evaluating}
Feiyang Yu, Mark Endo, Rayan Krishnan, Ian Pan, Andy Tsai, Eduardo~Pontes Reis, Eduardo Kaiser Ururahy~Nunes Fonseca, Henrique Min~Ho Lee, Zahra Shakeri~Hossein Abad, Andrew~Y. Ng, Curtis~P. Langlotz, Vasantha~Kumar Venugopal, and Pranav Rajpurkar.
\newblock Evaluating progress in automatic chest \uppercase{X}-ray radiology report generation.
\newblock {\em Patterns}, 4(9):100802, 2023.

\bibitem{zio2}
Enrico Zio.
\newblock {\em {The Monte Carlo Simulation Method for System Reliability and Risk Analysis}}.
\newblock Number 978-1-4471-4588-2 in Springer Series in Reliability Engineering. Springer, September 2013.

\bibitem{brito2022statistical}
Éder S.~Brito, Vera~L.D. Tomazella, and Paulo~H. Ferreira.
\newblock Statistical modeling and reliability analysis of multiple repairable systems with dependent failure times under perfect repair.
\newblock {\em Reliability Engineering \& System Safety}, 222:108375, 2022.

\end{thebibliography}

% Bibliography

\newpage
\appendix

\section*{Supporting Information}
\addcontentsline{toc}{section}{Supporting Information} % Adds the section to the Table of Contents

\section{Back to Basics: Reliability Engineering 101 for AI Systems}
\label{bathtub}

Reliability engineering is a well-established discipline in traditional engineering fields, focusing on ensuring that systems perform their intended function without failure for a specified period under stated conditions. The application of reliability engineering principles such as the bathtub curve, Mean Time to Failure (MTTF), and Mean Time Between Failures (MTBF) to AI systems is relatively novel. Insights into condition monitoring systems and the importance of sensor systems with multi-sensor fusion alongside AI systems highlight the potential for these engineering concepts to anticipate and manage AI system failures effectively \cite{jan2022, SURUCU2023119738, zohaib2023}. 

The integration of human reliability analysis (HRA) in AI systems, as discussed in the article and supported by incidentdatabase.ai, emphasizes the critical role of human factors in ensuring AI reliability. Recognizing human factors is crucial considering that AI systems are designed, operated, and interpreted by humans. The literature suggests that a comprehensive understanding of human-AI interaction is essential for developing more reliable and trustworthy AI systems.

In non-AI systems, the expectation is that the system performs consistently and its performance does not deteriorate or fail within certain bounds. In contrast, AI systems require not only consistent base-level performance, but also an expectation of improving performance or at least adaptability to incoming data. This makes reliability more challenging. Thus, the concept of resilience mathematics becomes important, similar to reliability, to measure performance improvement \cite{bilal1, bilal2}.

\subsection{Data-Driven Risk Management}

The proposed methodology integrates data-driven risk management with reliability engineering, offering a novel approach to AI safety. The literature supports the effectiveness of data-driven strategies in identifying and mitigating risks in complex systems. There are potential benefits in using big data and analytics for risk assessment and management in AI systems, suggesting that a data-driven approach can significantly enhance the ability to predict and prevent AI failures. Recent studies emphasize the necessity of integrating stringent safety guarantees in AI systems, especially those with high autonomy or used in safety-critical environments, to prevent harmful behaviors through robust quantitative safety frameworks. Additionally, the importance of regulatory compliance \cite{aldoseri2023, lazar2023}, such as adhering to GDPR guidelines \cite{cnil2024}, is underscored as a critical element that complements technical safety measures, ensuring AI systems respect individual rights and maintain trustworthiness across diverse applications \cite{dalrymple2024, bajcsy2024, lazar2023}.

In the transition from model development to deployment, it is crucial to bridge the gap between theoretical reliability and real-world performance of AI systems. This phase often reveals discrepancies that can be critical for system safety. To address this, we propose a continuous monitoring framework that begins in the development phase and extends through initial deployment into routine operational settings. This approach utilizes a combination of batch and real-time data retraining to adapt and refine the AI model continuously. By comparing the model's performance in production environments directly against its training performance, developers can identify and mitigate degradation or deviation from expected outcomes. This lifecycle perspective ensures that the AI system not only meets initial reliability standards but also maintains these standards in a dynamic real-world context, accommodating changes in data inputs, user interactions, and external conditions. This ongoing evaluation becomes a cornerstone for a data-driven, human-centered regulatory framework that enhances both the safety and trustworthiness of AI sytems.

Incorporating human developers, deployers, and end users as distinct elements within AI technical systems addresses critical challenges of performance deterioration and over-reliance that can emerge in production. As AI systems plateau or degrade over time, the interplay of human reliability and machine efficiency becomes crucial. Proactively designed alerts for both human and machine errors, along with clearly defined operational boundaries, can prevent biases and improve decision-making processes. This human-centric approach to AI systems, focusing on optimizing the symbiosis between human operators and AI, forms the basis of a robust reliability engineering strategy.

\subsection{The bathtub curve}

Reliability engineering uses the ``bathtub curve'' to model the lifecycle of systems, highlighting three phases of failure rates: early failures, random failures, and wear-out failures. This concept is directly applicable to AI systems. During the ``model design and development'' phase—similar to the design and testing of a new car—issues such as scope creep, design flaws, and partner disagreements often cause high initial failure rates, akin to the ``early failures'' phase in the bathtub curve. Once the AI model is refined and deployed, it enters the ``model deployment and operation'' phase, comparable to a new car bought by a consumer. Here, the system experiences more stable performance with occasional, unpredictable issues—this is the ``random failure'' phase. Finally, as the AI model ages and its algorithms or data become outdated, it may enter the ``wear-out'' phase, where performance gradually declines, necessitating updates or retraining.

\begin{figure}[htbp]
    \centering % Center the entire figure
    % Adjust the width of each subfigure to less than half the text width to fit side by side
    \begin{subfigure}[t]{0.45\textwidth}
        \centering % Center the subfigure
        \includegraphics[width=\textwidth]{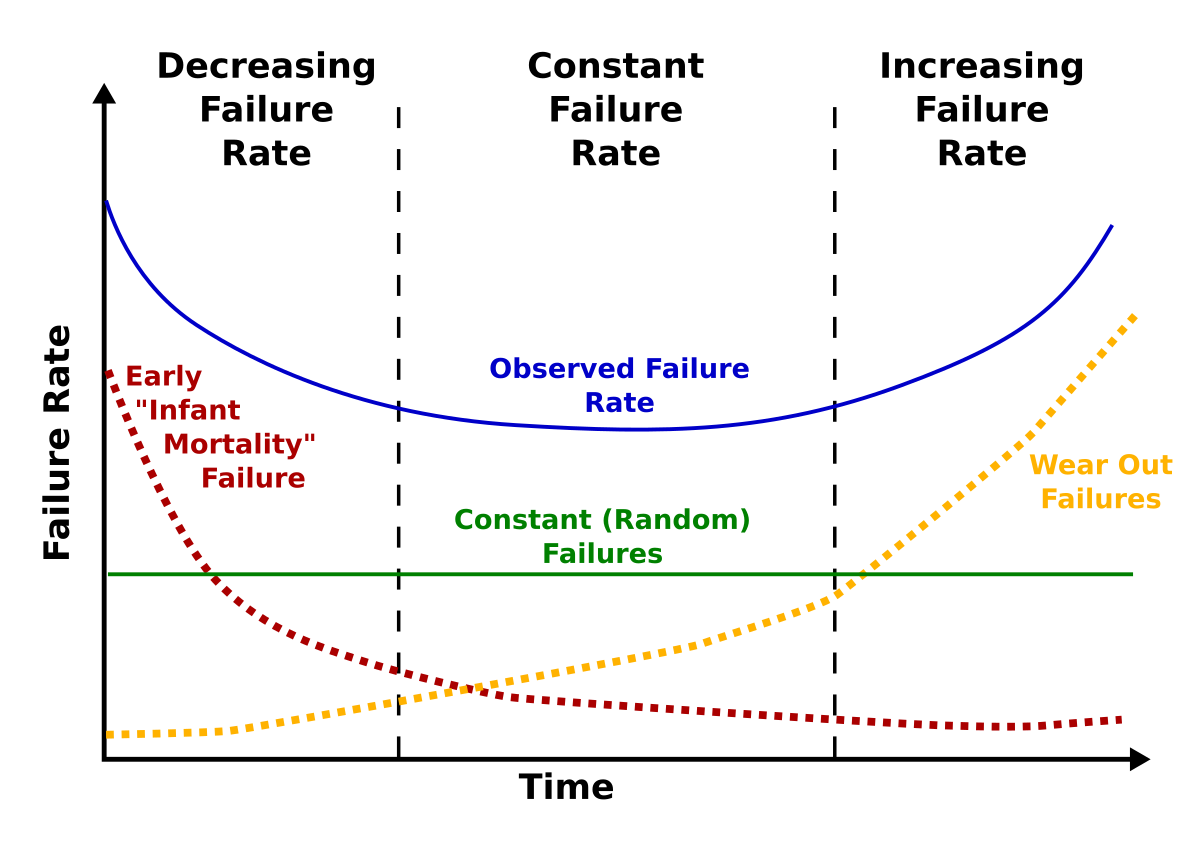}
        \caption{The `bathtub curve' failure function as a risk rate in AI systems. The curve combines a decreasing failure of early failure (red dotted line) and an increasing failure of wear-out failure (yellow dotted line), plus some constant failure of random failure (green, lower solid line).}
        \label{fig:bathtub}
    \end{subfigure}
    \hfill % Add horizontal space between subfigures
    \begin{subfigure}[t]{0.45\textwidth}
        \centering % Center the subfigure
        \includegraphics[width=\textwidth]{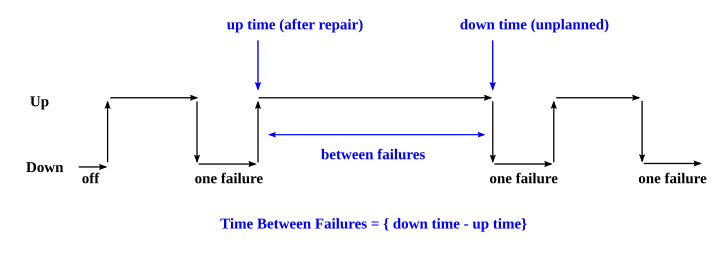}
        \caption{Time to and between various failure events, up-time and down-time.}
        \label{fig:timebetweenfailure}
        \caption*{\footnotesize \textbf{Notes}: MTBF is the "up-time" between two failure states of a repairable system during operation. The MTTF metric is associated with non-repairable components, whereas the MTBF metric is related to repairable components.}
    \end{subfigure}
    \caption{Risk Rates and Failure Times in AI Systems}
    \label{fig:risk_and_failure}
\end{figure}

\newpage 

\section{Probability Distributions in Reliability Engineering and Applications in AI Systems}
\label{appendix:probabilitydis}

Reliability engineering uses probability distributions to model the time to failure, error rates, and the overall system reliability. In AI systems, these distributions help model various aspects such as error propagation, system failures, and decision-making accuracy.

\subsection{Exponential Distribution}
The Exponential distribution is widely used in reliability engineering to model the time between failures in systems with a constant failure rate. This distribution is particularly relevant for AI systems that operate under continuous usage conditions.

\begin{equation}
f(t) = \lambda e^{-\lambda t}
\end{equation}
where \( \lambda \) is the failure rate, and \( t \) is the time. In the context of AI, \( \lambda \) can represent the rate of model errors or system breakdowns over time.

\subsection{Weibull Distribution}
The Weibull distribution models time-to-failure data where the failure rate is not constant but varies over time. This distribution is crucial for AI systems whose performance degrades due to factors like model drift or hardware aging.

\begin{equation}
f(t) = \frac{\beta}{\eta}\left(\frac{t}{\eta}\right)^{\beta-1}e^{-(t/\eta)^\beta}
\end{equation}
where \( \beta \) is the shape parameter indicating the nature of the failure rate (increasing, constant, or decreasing), and \( \eta \) is the scale parameter. This model is applicable in AI for analyzing systems where the failure rate increases as the system "ages" or learns over time.

For AI systems where failures may recur over time, the Non-Homogeneous Poisson Process (NHPP) is an effective model. The event intensity function is given by:
\[
k(t; \theta) = \frac{\beta}{\eta}\left(\frac{t}{\eta}\right)^{\beta-1}
\]
which allows for modeling the rate of recurrent failures as a function of time \cite{hong2023statistical}.

Out-of-distribution (OOD) detection is critical for AI reliability, as AI systems may encounter data that deviates significantly from the training set. As noted by \cite{hong2023statistical}, effective OOD detection can prevent AI system failures by identifying and appropriately handling these anomalies.

Adversarial attacks pose significant threats to AI reliability by subtly altering inputs to produce incorrect outputs. According to \cite{hong2023statistical}, these attacks can lead to misclassifications, which compromise the reliability and safety of AI systems. Methods to detect and mitigate adversarial attacks are therefore essential.

Effective test planning is essential for demonstrating AI reliability. The SMART framework advocates for structured test planning that includes the design of experiments and accelerated testing to assess AI performance under various conditions \cite{hong2023statistical}.

Degradation data models are used to monitor the performance deterioration of AI systems over time. For example, the general path model (GPM) can be applied, where the degradation path is modeled as:
\[
y(t) = D(t; \alpha) + \epsilon(t)
\]
allowing for the estimation of the time until the system performance falls below a critical threshold \cite{hong2023statistical}.

Accelerated testing is a valuable tool for quickly gathering reliability data on AI systems that are expected to have long operational lifetimes. By subjecting AI systems to intensified conditions, we can obtain reliability information more efficiently \cite{hong2023statistical}.

\subsection{Normal Distribution}
The Normal distribution is used in reliability engineering to model data that clusters around a mean, which is common in AI performance metrics, such as accuracy or precision scores.

\begin{equation}
f(x) = \frac{1}{\sigma \sqrt{2\pi}} e^{-\frac{1}{2} \left(\frac{x - \mu}{\sigma}\right)^2}
\end{equation}
where \( \mu \) is the mean, and \( \sigma \) is the standard deviation. In AI, this can model the distribution of prediction errors, assuming the errors follow a normal distribution.

\subsection{Log-Normal Distribution}
In cases where AI system processes follow multiplicative rather than additive processes, the Log-Normal distribution is more appropriate.

\begin{equation}
f(x) = \frac{1}{x\sigma\sqrt{2\pi}} e^{-\frac{(\ln x - \mu)^2}{2\sigma^2}}
\end{equation}
where \( \mu \) and \( \sigma \) are the parameters of the underlying normal distribution. This distribution is useful in AI systems to model phenomena such as the exponential growth of errors or accumulation of minor faults over time.

\subsection{Gamma Distribution}
The Gamma distribution is useful in modeling the time until an AI system experiences a certain number of failures.

\begin{equation}
f(t) = \frac{\lambda^k t^{k-1} e^{-\lambda t}}{(k-1)!}
\end{equation}
where \( k \) is the shape parameter representing the number of failures, and \( \lambda \) is the rate parameter. This is particularly useful in predicting when an AI system might need maintenance or retraining.

\subsection{Binomial Distribution}
The Binomial distribution models the number of successes in a series of independent experiments, such as AI model predictions.

\begin{equation}
P(X = k) = \binom{n}{k} p^k (1-p)^{n-k}
\end{equation}
where \( n \) is the number of trials, \( k \) is the number of successes, and \( p \) is the probability of success. This distribution can be used to model the success rate of an AI system over multiple runs or tests.

\newpage
\section{Application in AI System Reliability}
These distributions are fundamental in both reliability engineering and AI system analysis. By modeling the failure rates, error distributions, and success probabilities, engineers can better understand and predict AI system behavior, ultimately leading to more reliable and robust AI implementations.

\textbf{Characteristics of Discrete Failure or Improvement Events}

\begin{itemize}

\item Identifiability: Each failure event can be distinctly identified and documented. This includes knowing when the event occurred, under what conditions, and the specific nature of the failure.

\item Countability: Because these events are discrete, they can be counted. This allows for statistical analysis, such as calculating the failure rate over time.

\item Isolatability: Failure discrete events can often be isolated to specific components or aspects of the AI system, making it easier to diagnose and address the root cause.

\item Reproducibility: In some cases, if the conditions leading to the failure are well understood, the failure event can be reproduced. This is particularly useful for testing and verification purposes.

\end{itemize}

\textbf{Importance in AI System Reliability}

Understanding and managing failure discrete events are critical for several reasons:

\begin{itemize}
\item Improving System Reliability: By analyzing these events, engineers can identify patterns or common causes of failure, leading to targeted improvements in the AI system’s design, coding, or operational procedures.

\item Risk Management: Identifying potential failure modes and their impacts helps in developing strategies to mitigate these risks, such as implementing redundancy, improving error handling, or enhancing system monitoring.

\item Maintenance and Repair: Discrete failure events provide clear signals that maintenance or repair actions are needed, allowing for timely interventions to restore system functionality.

\item Compliance and Safety: In regulated industries, understanding failure modes and their consequences is essential for compliance with safety standards and regulations.
\end{itemize}

\newpage 

\section{Accelerated Life Testing for AI Systems}
\label{sec:ALT}
\subsection{Introduction to Accelerated Life Testing (ALT)}

Accelerated Life Testing (ALT) is a critical methodology in reliability engineering, widely used to predict the life expectancy of complex engineering systems by subjecting them to higher-than-normal stress levels to induce failures more quickly. This approach is commonly applied in industries such as aerospace, automotive, and electronics, where components and systems are exposed to conditions such as elevated temperatures, increased vibration, or higher loads to simulate years of operation within a shorter period. For instance, in the aerospace industry, ALT is utilized to evaluate the fatigue life of aircraft components under cyclic loading, where the stress levels are significantly higher than during normal flight conditions \cite{relevant_reference_1}.

The principle of ALT involves assuming a relationship between the stress applied and the lifespan of the system. This relationship is often modeled using parametric life distributions like the Weibull or Lognormal distribution. The goal is to extrapolate the life of the system under normal operating conditions based on the accelerated testing data. This method allows for faster product development cycles, reduced costs, and early identification of potential failure modes, thereby ensuring product reliability.

When comparing ALT in traditional engineering systems to AI systems, particularly those involving models like Large Language Models (LLMs), several similarities and differences emerge. Traditional engineering systems undergo physical stress, while AI systems face computational and data-related stress. For AI systems, "stress" could be defined as factors that exacerbate model performance degradation, such as increasing data noise, introducing adversarial examples, or simulating edge cases in operational environments. These stresses can affect the accuracy and reliability of AI models, similar to how physical stress impacts the lifespan of mechanical components.

\subsection{Mathematical Framework for ALT in AI Systems}

Accelerated Life Testing traditionally relies on a stress-life relationship, where the life \( T \) of a system under a given stress level \( S \) is modeled using a parametric distribution. The most common models used are the Weibull and Lognormal distributions. The life distribution under stress \( S \) can be described by the following general form:

\[
\eta(S) = \eta_0 \cdot \exp(\beta \cdot S)
\]

where:
\begin{itemize}
    \item \( \eta(S) \) is the scale parameter of the life distribution under stress \( S \).
    \item \( \eta_0 \) is the scale parameter at the nominal stress level \( S_0 \).
    \item \( \beta \) is the stress-life coefficient, which quantifies how the applied stress influences the system's life.
\end{itemize}

The corresponding probability density function (PDF) of the life distribution under stress \( S \) is:

\[
f(T|S) = \frac{\beta}{\eta(S)} \left(\frac{T}{\eta(S)}\right)^{\beta - 1} \exp\left[-\left(\frac{T}{\eta(S)}\right)^\beta\right]
\]

This equation is particularly useful in estimating the reliability \( R(t|S) \) of the system over time \( t \) under the stress \( S \):

\[
R(t|S) = \exp\left[-\left(\frac{t}{\eta(S)}\right)^\beta\right]
\]

This reliability function describes the probability that a system will perform its intended function without failure up to time \( t \) under stress level \( S \).

\subsection{Application to AI Systems}

In the context of AI systems, we define ''stress'' as factors that impact the model's performance and accuracy. These factors can include data quality, adversarial inputs, and processing loads. The stress-life relationship for AI models can be analogously defined using a similar framework:

\[
\lambda(S) = \lambda_0 \cdot \exp(\gamma \cdot S)
\]

where:
\begin{itemize}
    \item \( \lambda(S) \) is the failure rate of the AI model under stress \( S \).
    \item \( \lambda_0 \) is the baseline failure rate under nominal conditions.
    \item \( \gamma \) is the stress sensitivity coefficient, reflecting the AI model's sensitivity to the stress factor.
\end{itemize}

The reliability of the AI system under stress \( S \) over time \( t \) can be expressed as:

\[
R(t|S) = \exp\left[-\lambda(S) \cdot t\right]
\]

This equation captures the probability that an AI system will continue to perform without failure up to time \( t \) under the influence of the stress \( S \).

\subsection{Time-Dependent Reliability Analysis}

Time-dependent reliability analysis extends the above concepts to account for the fact that the failure rate \( \lambda(t|S) \) may vary over time as the system or model ages or as operational conditions change. The cumulative distribution function (CDF) of life \( T \) under time-dependent stress can be modeled as:

\[
F_T(t) = 1 - \exp\left(-\int_0^t \lambda(\tau|S) d\tau\right)
\]

This approach is particularly useful in scenarios where the stress factors themselves may change over time, such as a gradually increasing data noise level or an escalating complexity in tasks performed by an AI model.

The figure below illustrates the stress-life relationship obtained from time-dependent reliability analysis. As shown in the figure, the life distributions at different stress levels are connected through the underlying physics model. This connection is essential for guiding data-based ALT design by fusing information from both experimental data and physics-based models \cite{relevant_reference_2}.

\begin{figure}[htbp]
    \centering
    \includegraphics[width=0.6\textwidth]{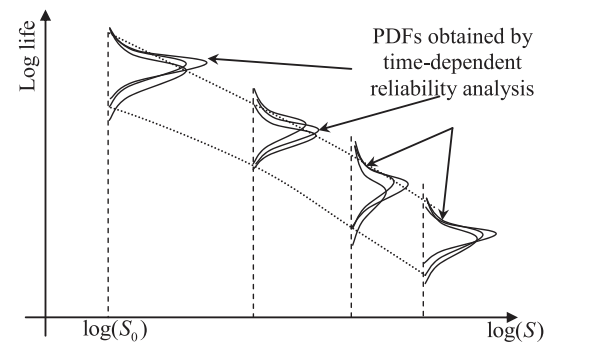}
    \caption{Illustration of stress-life relationship obtained from time-dependent reliability analysis \cite{relevant_reference_1}.}
    \label{fig:ALT}
\end{figure}

\subsection{Bayesian Updating for Reliability Estimation}

In practical applications of ALT, particularly for AI systems, it is crucial to incorporate uncertainty due to limited data or model imperfections. Bayesian methods are often used to update the reliability estimates as new data becomes available during testing. The updated reliability estimate \( R(t|S, \theta) \), where \( \theta \) represents the parameters of the life distribution (such as the shape and scale parameters in a Weibull distribution), can be expressed as:

\[
R(t|S, \theta) = \int_\Theta R(t|S, \theta) \cdot p(\theta|D) \, d\theta
\]

where:
\begin{itemize}
    \item \( \Theta \) represents the parameter space.
    \item \( p(\theta|D) \) is the posterior distribution of \( \theta \) given the observed data \( D \).
\end{itemize}

Bayesian updating is particularly powerful in AI system testing, where the initial model parameters may be uncertain due to the complexity of the model or the novelty of the application domain. As more data is gathered, the reliability estimates become more precise, reducing epistemic uncertainty and improving the overall confidence in the AI system's performance.

\subsection{Surrogate Modeling for Efficiency}

Given the computational intensity of full-scale reliability analysis, especially when integrating time-dependent factors and Bayesian updating, surrogate models are often employed. Surrogate models approximate the complex relationships between stress, time, and reliability, allowing for faster computations. These models are typically constructed using methods such as Gaussian processes or Kriging, which can efficiently interpolate the results of detailed simulations.

The general form of a Kriging surrogate model for reliability estimation is:

\[
\hat{R}(t|S) = h(t, S)^\top \beta + Z(t, S)
\]

where:
\begin{itemize}
    \item \( h(t, S) \) is a vector of basis functions representing the trend.
    \item \( \beta \) is a vector of coefficients.
    \item \( Z(t, S) \) is a Gaussian process representing the residual between the trend and the actual reliability.
\end{itemize}

This approach allows for the rapid evaluation of reliability across different stress scenarios, facilitating the design and optimization of ALT plans for AI systems.

\newpage 
\section{Examples of AI System Failures}
\label{appendixanalysis}

In this appendix, we provide illustrative examples of AI system failures to highlight how human errors and technical breakdowns can occur in real-world applications. These examples serve to underscore the importance of reliability engineering in AI systems and the role of human factors in system failures.

\subsection{Resources and Datasets}

We have identified few resources that catalog AI incidents across various industries and applications:

\begin{itemize} 
\item \textbf{AI Incident Database}: \url{https://incidentdatabase.ai/apps/incidents/} \ A publicly available repository that collects and documents AI-related incidents, providing insights into how AI systems can fail and the impact of those failures on society.
\item \textbf{MIT AI Risk Repository}: \url{https://airisk.mit.edu/} \\
A comprehensive database curated by MIT that focuses on the risks associated with AI technologies, aiming to inform stakeholders about potential hazards and mitigation strategies.

\item \textbf{OpenAI Status Page}: \url{https://status.openai.com/} \\
An official status page that provides real-time updates on the operational status of OpenAI's services, including any incidents or outages that may affect users.

\end{itemize}

\subsection{Examples of AI System Failures}

Table \ref{incidentstable} presents a selection of AI incidents, illustrating how failures can occur due to both machine and human errors. These examples are drawn from the AI Incident Database and the MIT AI Risk Repository, structured to provide insights into the root causes and potential preventative measures. Table \ref{smalltable} provides a small sample of the AI incidents reported.

\begin{table}[htbp]
\centering
\caption{Sample of Incidents Categorized by Human Failure and Technical Implications}
\label{smalltable}
\begin{tabular}{|>{\raggedright\arraybackslash}p{3cm}|
                >{\raggedright\arraybackslash}p{3cm}|
                >{\raggedright\arraybackslash}p{3cm}|
                >{\raggedright\arraybackslash}p{3cm}|}
\hline
\textbf{Incident} & \textbf{Type of Human Failure} & \textbf{Cost} & \textbf{Implications for AI System} \\ \hline
Facebook Algorithm Amplifies Misinformation & Misuse of algorithm for content amplification & Reputational damage, societal harm & Need for better contextual awareness and algorithm adjustments \\ \hline
AI-Generated Deepfake Scams & Exploitation of AI to create false content & Millions lost in fraud (e.g., \$25M from Arup) & Improved deepfake detection and authentication mechanisms \\ \hline
AI Facial Recognition Bias in Tax Audits & Algorithmic bias in tax audits & Legal costs, public trust issues & Model tuning, data cleaning, and human oversight needed \\ \hline
\end{tabular}
\end{table}

\begin{figure}[htbp]
    \centering
    \includegraphics[width=0.48\textwidth]{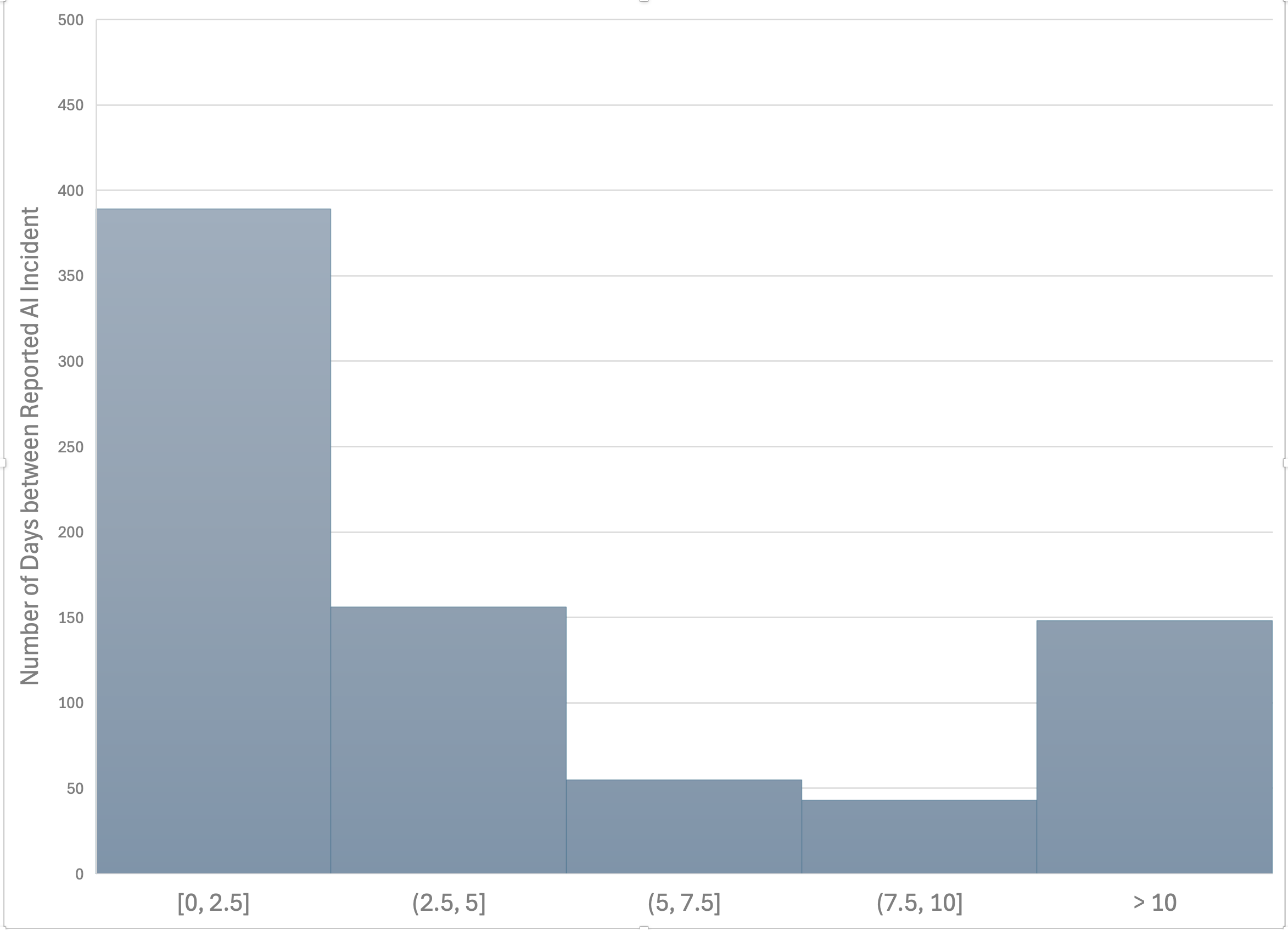}
    \caption{Histogram of Time Between Failures (TBF) for AI Incidents: Distribution of Days Between Reported Events.}
    \label{fig:tbf_histogram}
    \vspace{0.5em} % Add some space between the caption and the source
    \small\textit{Source:} \href{https://incidentdatabase.ai/apps/incidents/}{AI Incident Database}
\end{figure}

The histogram in Figure \ref{fig:tbf_histogram} shows the distribution of Time Between Failures (TBF) for the AI incidents dataset. The majority of incidents occur with a short TBF, as seen by the large number of events in the first bin, which covers the range from 0 to 2.5. This suggests that many AI systems experience failures or harmful incidents relatively soon after their deployment or between successive failures. The distribution reveals that AI system failures are relatively frequent in the short term, indicating the need for more robust engineering and testing practices to improve reliability.

\begin{figure}[htbp]
    \centering
    \includegraphics[width=0.4\textwidth]{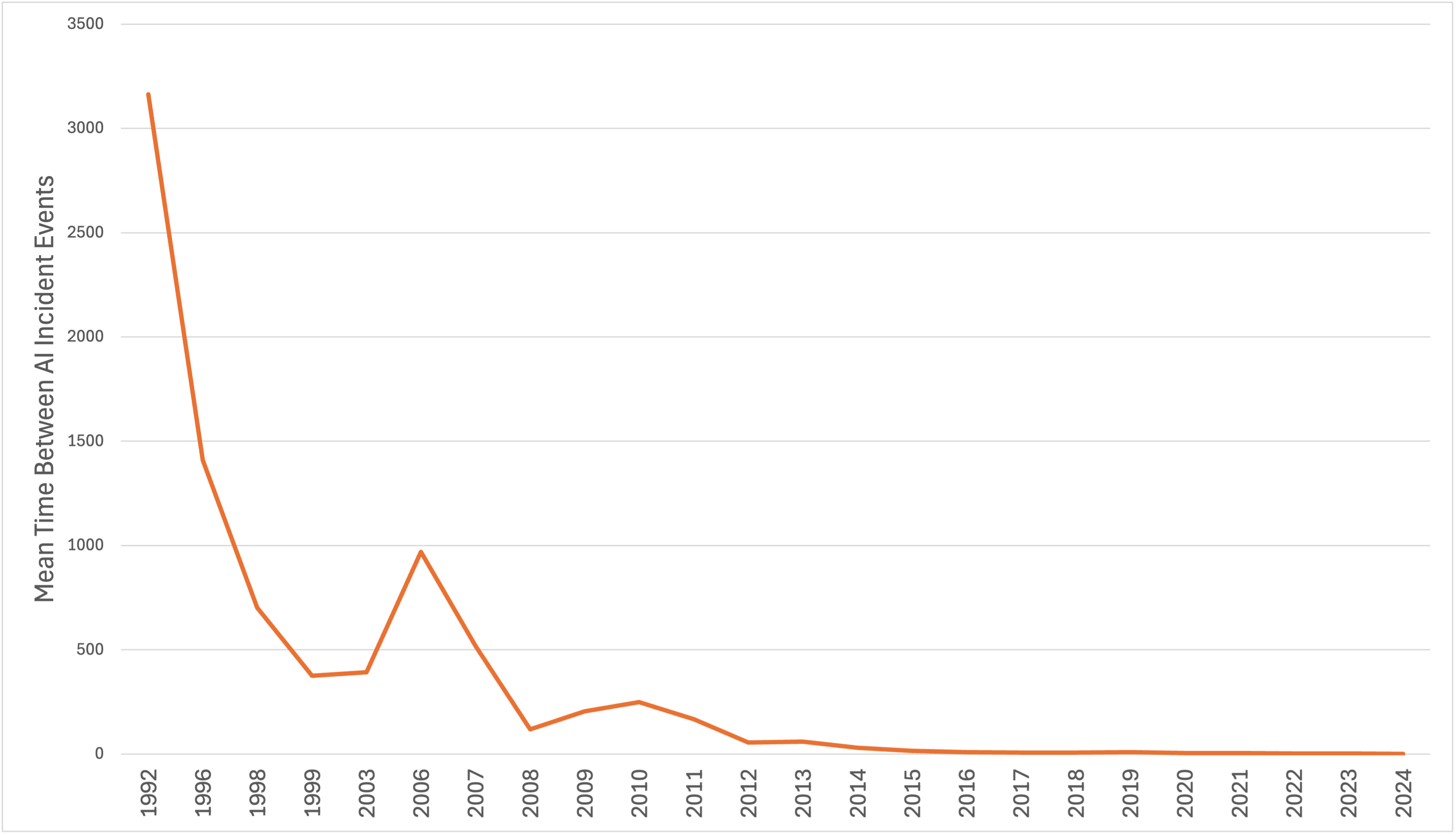}
    \caption{Mean Time Between Failures (MTBF) Over Time}
    \label{fig:mtbf_chart}
    \small\textit{Source:} \href{https://incidentdatabase.ai/apps/incidents/}{AI Incident Database}
\end{figure}

The Mean Time Between Failures (MTBF) chart, shown in Figure \ref{fig:mtbf_chart}, illustrates a clear downward trend in the number of days between AI-related incidents over time. Initially, there are long intervals between incidents in the early 1990s (3164 days in 1992), but the frequency of incidents increases sharply in subsequent years. By 2024, the MTBF drops to just 2 days. This downward trend suggests a higher frequency of failures, although reporting bias could also be a contributing factor.

\begin{figure}[htbp]
    \centering
    \includegraphics[width=0.5\textwidth]{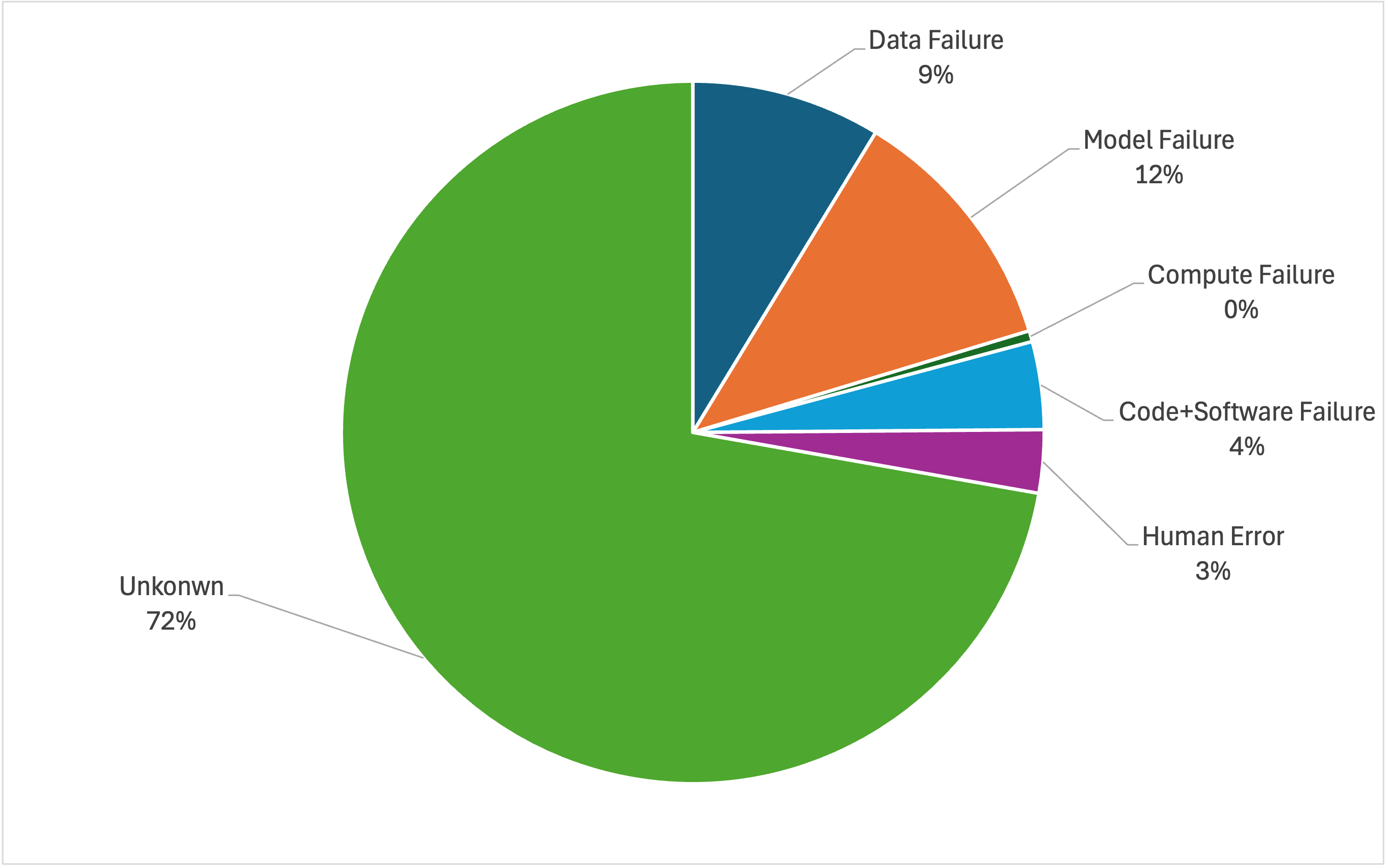}
    \caption{AI System Failure Modes Distribution}
    \label{fig:failure_pie_chart}
    \small\textit{Source:} \href{https://incidentdatabase.ai/apps/incidents/}{AI Incident Database}
\end{figure}

The pie chart in Figure \ref{fig:failure_pie_chart} illustrates the distribution of AI system failure modes across different categories. The largest category (72\%) is labeled as "Unknown," highlighting the need for deeper research and manual curation to accurately categorize each incident. "Model Failure" (92 cases) and "Data Failure" (69 cases) follow, while "Human Error" accounts for 23 incidents. These insights emphasize the critical need for more refined reporting mechanisms to better distinguish between technical and human causes in AI system incidents.

These examples are provided for illustrative purposes and represent a snapshot of incidents that have been publicly reported. The information available may be limited, and in some cases, the full technical details of the failures are not disclosed. Therefore, while these incidents highlight real-world failures that are often caught in production, they should be interpreted with caution, and further investigation is recommended for a comprehensive understanding.

\newpage 
\scriptsize % Further reduce font size for better fit
\begin{longtable}{|>{\raggedright\arraybackslash}p{1.3cm}|>{\raggedright\arraybackslash}p{1.1cm}|>{\raggedright\arraybackslash}p{1.1cm}|>{\raggedright\arraybackslash}p{3cm}|>{\raggedright\arraybackslash}p{1.5cm}|>{\raggedright\arraybackslash}p{1.5cm}|>{\raggedright\arraybackslash}p{1.8cm}|}
\caption{Small Sample of Subsystem and Component Level Failure Modes: Anthropic System} \label{anthropic_system_failures} \\
\hline
\textbf{Start Date} & \textbf{End Date} & \textbf{Incident Type} & \textbf{Incident Description} & \textbf{Failure Mode} & \textbf{Root Cause} & \textbf{Component} \\
\hline
\endfirsthead
\hline
\textbf{Start Date} & \textbf{End Date} & \textbf{Incident Type} & \textbf{Incident Description} & \textbf{Failure Mode} & \textbf{Root Cause} & \textbf{Component} \\
\hline
\endhead
\hline
\endfoot
\hline
\endlastfoot

10/30/2024 & 10/30/2024 & Elevated errors for Claude 3.5 & Elevated errors for requests to Claude 3.5 Sonnet. Success rates returned to normal. & Elevated Error & System Load Issue & Claude 3.5 Sonnet \\
\hline
10/28/2024 & 10/28/2024 & Analysis tool prevents viewing & Users unable to view Artifacts on Claude.ai due to a tool error. & Viewing Error & Tool Misconfiguration & Analysis Tool \\
\hline
10/27/2024 & 10/27/2024 & Login failures & Errors logging into Claude.ai. & Login Failure & Authentication Outage & Authentication Service \\
\hline
10/25/2024 & 10/25/2024 & Errors on Claude 3.5 AWS Bedrock & Errors on requests to Claude 3.5 Sonnet v2 on AWS Bedrock. & Elevated Error & Configuration Issue & Claude 3.5 Sonnet \\
\hline
... & ... & ... & ... & ... & ... & ... \\
\hline
\end{longtable}

\vspace{0.5cm} % Small space between tables for readability

% Second Table: GitHub System
\scriptsize
\begin{longtable}{|>{\raggedright\arraybackslash}p{1.3cm}|>{\raggedright\arraybackslash}p{1.1cm}|>{\raggedright\arraybackslash}p{1.2cm}|>{\raggedright\arraybackslash}p{2.8cm}|>{\raggedright\arraybackslash}p{1.5cm}|>{\raggedright\arraybackslash}p{1.5cm}|>{\raggedright\arraybackslash}p{1.8cm}|}
\caption{Small Sample of Subsystem and Component Level Failure Modes: GitHub System} \label{github_system_failures} \\
\hline
\textbf{Start Date} & \textbf{End Date} & \textbf{Incident Type} & \textbf{Incident Description} & \textbf{Failure Mode} & \textbf{Root Cause} & \textbf{Component} \\
\hline
\endfirsthead
\hline
\textbf{Start Date} & \textbf{End Date} & \textbf{Incident Type} & \textbf{Incident Description} & \textbf{Failure Mode} & \textbf{Root Cause} & \textbf{Component} \\
\hline
\endhead
\hline
\endfoot
\hline
\endlastfoot

10/30/2024 & 10/30/2024 & Incident with Actions & Resolved action run failures during peak hours & Workflow Failures & System configuration issue & Actions Service \\
\hline
10/24/2024 & 10/24/2024 & Incident with GitHub Community Discussions & 500 error due to invalid YAML template & User Access Error & Configuration error in YAML file & Community Service \\
\hline
10/11/2024 & 10/12/2024 & Disruption with some GitHub services & DNS issues due to failed database migration affecting multiple services & DNS Failure & Migration-related database issues & DNS Service \\
\hline
09/30/2024 & 09/30/2024 & Incident with Codespaces & Unable to create Codespaces in Central India due to storage constraints & Capacity Limit & Regional storage limitation & Codespaces \\
\hline
... & ... & ... & ... & ... & ... & ... \\
\hline
\end{longtable}

\vspace{0.5cm} % Small space between tables for readability

% Third Table: GCP System (modified to show only 4 incidents)
\scriptsize
\begin{longtable}{|>{\raggedright\arraybackslash}p{1.3cm}|>{\raggedright\arraybackslash}p{1.1cm}|>{\raggedright\arraybackslash}p{1.2cm}|>{\raggedright\arraybackslash}p{2.8cm}|>{\raggedright\arraybackslash}p{1.5cm}|>{\raggedright\arraybackslash}p{1.5cm}|>{\raggedright\arraybackslash}p{1.8cm}|}
\caption{Small Sample of Subsystem and Component Level Failure Modes: GitHub System} \label{gcp_system_failures} \\
\hline
\textbf{Start Date} & \textbf{End Date} & \textbf{Incident Type} & \textbf{Incident Description} & \textbf{Failure Mode} & \textbf{Root Cause} & \textbf{Component} \\
\hline
\endfirsthead
\hline
\textbf{Start Date} & \textbf{End Date} & \textbf{Incident Type} & \textbf{Incident Description} & \textbf{Failure Mode} & \textbf{Root Cause} & \textbf{Component} \\
\hline
\endhead
\hline
\endfoot
\hline
\endlastfoot

10/30/2024 & 10/30/2024 & Incident with Actions & Resolved action run failures during peak hours & Workflow Failures & System configuration issue & Actions Service \\
\hline
10/24/2024 & 10/24/2024 & Incident with GitHub Community Discussions & 500 error due to invalid YAML template & User Access Error & Configuration error in YAML file & Community Service \\
\hline
10/11/2024 & 10/12/2024 & Disruption with some GitHub services & DNS issues due to failed database migration affecting multiple services & DNS Failure & Migration-related database issues & DNS Service \\
\hline
09/30/2024 & 09/30/2024 & Incident with Codespaces & Unable to create Codespaces in Central India due to storage constraints & Capacity Limit & Regional storage limitation & Codespaces \\
\hline
... & ... & ... & ... & ... & ... & ... \\
\hline
\end{longtable}

\newpage

\setlength{\emergencystretch}{8em} % Increase as needed

% Increase row spacing
\renewcommand{\arraystretch}{1.3}

% Caption settings
\captionsetup{font=small}

% Set font size for the table to small to fit content better
\small

\begin{longtable}{|>{\raggedright\arraybackslash}p{2cm}|
                    >{\raggedright\arraybackslash}p{2.6cm}|
                    >{\raggedright\arraybackslash}p{1.8cm}|
                    >{\raggedright\arraybackslash}p{1.5cm}|
                    >{\raggedright\arraybackslash}p{4cm}|}
\caption{A Small Sample of AI Incidents and How Errors Could Have Been Mitigated} \label{incidentstable} \\
\hline
\textbf{Incident} & \textbf{Description} & \textbf{Error Type} & \textbf{Subsystem} & \textbf{Root Cause and Possible Fix} \\
\hline
\endfirsthead
\multicolumn{5}{c}{\tablename\ \thetable\ -- \textit{Continued from previous page}} \\
\hline
\textbf{Incident} & \textbf{Description} & \textbf{Error Type} & \textbf{Subsystem} & \textbf{Root Cause and Possible Fix} \\
\hline
\endhead
\hline
\multicolumn{5}{|r|}{\textit{Continued on next page}} \\
\hline
\endfoot
\hline
\endlastfoot

Las Vegas Bus Accident & Self-driving shuttle collided with truck on its first day. & Machine + Human Error & Model, Human & Inadequate sensor calibration for human-driven cases. \newline \textbf{Fix:} Improved sensor calibration and simulation testing. \\
\hline
Uber AV Accident & Uber AV killed pedestrian in Arizona. & Machine + Human Error & Model, Human & Weak emergency algorithms for real-world situations. \newline \textbf{Fix:} Emergency braking and safety protocols. \\
\hline
YouTube Kids Filtering & Inadequate content filtering allowed disturbing videos for children. & Machine Error & Data, Algorithm & Lack of comprehensive filters during training. \newline \textbf{Fix:} Stricter content moderation filters. \\
\hline
Bias in NLP Embeddings & Gender bias found in popular NLP embeddings. & Machine Error & Model & Inherent bias in training data. \newline \textbf{Fix:} Fairness-aware training methods. \\
\hline
Northpointe Risk Model Bias & Penal system model biased against Black people. & Machine Error & Data, Model & Insufficient bias correction in model training. \newline \textbf{Fix:} Regular audits and fairness metrics. \\
\hline
Robotic Surgery Errors & Malfunctions in robotic surgeries, leading to injuries. & Machine + Human Error & Hardware, Model & Inadequate testing between hardware and software. \newline \textbf{Fix:} Rigorous integration testing and safety checks. \\
\hline
Tesla Autopilot Crashes & Accidents occurred while Tesla’s autopilot was in use. & Machine Error & Model, Hardware & Insufficient real-time monitoring. \newline \textbf{Fix:} Enhanced real-time safety measures. \\
\hline
VW Robot Kills Worker & Worker killed by robot at Volkswagen plant. & Machine Error & Hardware & Lack of human-safety interaction protocols. \newline \textbf{Fix:} Safety protocols and fail-safes. \\
\hline
Security Robot Incident & Security robot collided with child, causing injury. & Machine Error & Model, Hardware & Inadequate proximity sensing. \newline \textbf{Fix:} Improved safety algorithms and sensors. \\
\hline
AI Misinformation Spread & AI model repeated disinformation. & Machine Error & Data, Model & Inadequate filtering of misinformation. \newline \textbf{Fix:} Better filtering and fact-checking algorithms. \\
\hline
AI-Assisted Fraud & AI used to create fraud schemes. & Machine Error & Model, Algorithm & Weak adversarial defenses. \newline \textbf{Fix:} Improved fraud detection and defenses. \\
\hline
Image Classification Bias & Google Photos misclassified Black people as gorillas. & Machine Error & Model, Data & Bias in image classification training data. \newline \textbf{Fix:} Debiasing and fairness checks. \\
\hline
DWP Fraud Detection Errors & Algorithm wrongly flagged 200,000 benefit claims. & Machine Error & Data, Model & Over-sensitive fraud detection thresholds. \newline \textbf{Fix:} Adjust thresholds and improve accuracy checks. \\
\hline
Racial Bias in Image Search & Racially biased image search results on Google. & Machine Error & Data, Model & Bias in training data. \newline \textbf{Fix:} Bias audits and balanced datasets. \\
\hline
LLM in Scams & LLMs used to create malware and phishing scams. & Human Error & Model, Human & Lack of security in training data. \newline \textbf{Fix:} Security-focused datasets and adversarial training. \\
\hline
Biased Sentiment Analysis & NLP API showed biased sentiment analysis results. & Machine Error & Model, Data & Bias in training and evaluation data. \newline \textbf{Fix:} Fairness evaluation in model development. \\
\hline
Kamala Harris Deepfake & Deepfake of Kamala Harris shared online. & Human Error & Model, Human & Lack of deepfake detection. \newline \textbf{Fix:} Develop deepfake detection algorithms. \\
\hline
Samsung Data Leak & Engineers accidentally leaked data via ChatGPT. & Human Error & Human, Model & Lack of data privacy protocols. \newline \textbf{Fix:} Strengthen data protection policies. \\
\hline
ESPN Misses Alex Morgan Event & AI-generated recap missed Alex Morgan’s retirement. & Machine Error & Model, Algorithm & Poor event recognition in personalization algorithms. \newline \textbf{Fix:} Quality checks on personalization algorithms. \\

\end{longtable}

\newpage

\begin{longtable}{|>{\raggedright\arraybackslash}p{1.9cm}|
                    >{\raggedright\arraybackslash}p{2.8cm}|
                    >{\raggedright\arraybackslash}p{3.5cm}|
                    >{\raggedright\arraybackslash}p{3.5cm}|}
\caption{Critical AI Software Subsystems with Common Failure Types and Metrics} \\
\hline
\textbf{Name} & \textbf{Description} & \textbf{Common Failure Types} & \textbf{Metrics and Measures} \\
\hline
\endfirsthead
\hline
\textbf{Name} & \textbf{Description} & \textbf{Common Failure Types} & \textbf{Metrics and Measures} \\
\hline
\endhead
\hline
\multicolumn{4}{|r|}{\textit{Continued on next page}} \\
\hline
\endfoot
\hline
\endlastfoot

Snowflake & A cloud data platform for data warehousing and analytics & Data drift, Data corruption & Data quality audits, real-time monitoring \\
\hline
Databricks & Unified analytics platform for big data and machine learning & Incomplete data, Biased samples & Data diversity indexes, entropy measures \\
\hline
Google BigQuery & A fully-managed, serverless data warehouse & Incorrect transformation, Data integrity issues & Statistical analysis, data sampling \\
\hline
AWS Redshift & A fully-managed data warehouse service in the cloud & Data security breaches, Data governance issues & Security audits, compliance checks \\
\hline
Elasticsearch & A distributed search and analytics engine & Query performance issues, Index corruption & Query latency, index health checks \\
\hline
TensorFlow & Open-source library for machine learning and AI & Overfitting, Underfitting & Model accuracy, validation scores \\
\hline
PyTorch & An open-source machine learning library based on the Torch library & Model drift, Misinterpretation of inputs & Drift detection metrics, input validation \\
\hline
Scikit-learn & A machine learning library for Python & Poor generalization, High error rates & Generalization measures, error rates \\
\hline
OpenAI (GPT) & Large language model for generating human-like text & Ethical issues, Misuse & Ethical compliance checks, usage audits \\
\hline
Mistral & A high-performance large language model & Performance degradation, Resource exhaustion & Performance benchmarks, resource utilization \\
\hline
LLaMA & Meta's language model for various NLP tasks & Scalability issues, Maintenance complexity & Scalability metrics, maintenance logs \\
\hline
AWS EC2 & Scalable computing capacity in the Amazon Web Services cloud & Computational bottlenecks, Hardware failures & Uptime, MTBF (Mean Time Between Failures) \\
\hline
Google Cloud Platform (GCP) & Comprehensive suite of cloud computing services & Latency, Delayed responses & Response time, load balancing efficiency \\
\hline
Microsoft Azure & Cloud computing services for building, testing, deploying, and managing applications & Resource scaling issues, Reliability problems & Resource utilization, reliability metrics \\
\hline
IBM Cloud & Cloud platform offering IaaS, SaaS, and PaaS & Downtime, Performance inconsistencies & Service uptime, performance logs \\
\hline
Jupyter Notebooks & An open-source web application for creating and sharing documents & Execution errors, Compatibility issues & Execution success rates, compatibility checks \\
\hline
VSCode & A source-code editor developed by Microsoft & Code integration issues, Debugging difficulties & Code review coverage, debugging time \\
\hline
GitHub & A platform for version control and collaboration & Merge conflicts, Versioning errors & Merge conflict rates, version control logs \\
\hline
Kubernetes & An open-source system for automating deployment, scaling, and management of containerized applications & Orchestration failures, Resource misallocation & Orchestration success rates, resource allocation metrics \\
\hline
Microservice Architecture & Architectural style that structures an application as a collection of loosely coupled services & Service communication failures, Deployment issues & Service uptime, deployment success rates \\
\hline
RAGAS & Framework for Initial LLM & Context Misalignment, Response Deviation & Context Precision, Rubrics-based Scoring \\
\hline
ARES & Adaptive Evaluation & Continuous Feedback Loop Failures & MRR, NDCG \\
\hline
TraceLoop & Full System Tracing & Information Inconsistencies, Lack of Provenance & Information Gain, Factual Consistency \\
\hline
Galileo & Enterprise Integration & Poor Context Adherence, Scalability Issues & Custom Metrics, Integration Success Rates \\
\hline
TruLens & Domain-specific Optimization & Domain-specific Drift, Contextual Errors & Domain-specific Accuracy, Precision Metrics \\
\end{longtable}

\end{document}